%% file: main.tex
\documentclass{article}

\usepackage{times}
\usepackage{iclr2026/iclr2026_conference}

\input{config.tex}

\acrodef{VLM}{Vision-Language Model}
\acrodef{MVR}{Multi-view Video Reward Shaping}

\definecolor{gblue}{HTML}{4285F4}

\title{MVR: Multi-view Video Reward Shaping for Reinforcement Learning}

\author{%
Lirui Luo\textsuperscript{1,2}\thanks{Equal contribution.},
Guoxi Zhang\textsuperscript{2}\footnotemark[1],
Hongming Xu\textsuperscript{2},
Yaodong Yang\textsuperscript{1,2},
Cong Fang\textsuperscript{1}\thanks{Corresponding authors.},
Qing Li\textsuperscript{2}\footnotemark[2]
\\
\normalfont
\textsuperscript{1} School of Intelligence Science and Technology, Peking University \\
\textsuperscript{2} State Key Laboratory of General Artificial Intelligence, BIGAI \\
\texttt{fangcong@pku.edu.cn, dylan.liqing@gmail.com} \\
Project page: \url{https://mvr-rl.github.io/}
}

\iclrfinalcopy % Camera-ready (non-anonymous)

\begin{document}

\maketitle

\begin{abstract}
\input{0.abstract}
\end{abstract}

\input{1.introduction}
\input{2.related-work}
\input{3.problem_statement}
\input{4.method}

\input{5.experiments}

\input{6.conclusions}

\section*{Acknowledgments}
C.~Fang was supported by the National Science and Technology Major Project (No.~2022ZD0114902).
% 中文标注：新一代人工智能国家科技重大专项（编号：2022ZD0114902）

\bibliographystyle{iclr2026/iclr2026_conference}
\bibliography{reference_header,references}

\appendix
\input{appendix}

\end{document}

%% file: config.tex
\usepackage[utf8]{inputenc} % allow utf-8 input
\usepackage[T1]{fontenc}    % use 8-bit T1 fonts
\usepackage{microtype,inconsolata}

\usepackage{algorithmic,algorithm}

\usepackage[utf8]{inputenc}

\usepackage{graphicx} \graphicspath{{figures/}}
\usepackage{booktabs,tabularx,colortbl,multirow,multicol,array,makecell,tabularray}
\usepackage{subcaption}
\usepackage{caption}

\usepackage[switch]{lineno}
\usepackage{threeparttable}

\usepackage{amsmath,amssymb,mathtools,amsthm,nicefrac}

\usepackage{acronym}
\usepackage{enumitem}
\usepackage{balance}
\usepackage{xspace}
\usepackage{setspace}

\usepackage[breaklinks,colorlinks=true,citecolor=blue,linkcolor=black,urlcolor=blue]{hyperref}
\usepackage[capitalise,noabbrev,nameinlink]{cleveref}
\usepackage{wrapfig}

\theoremstyle{plain}
\newtheorem{theorem}{Theorem}[section]

\theoremstyle{definition}
\newtheorem{definition}[theorem]{Definition}

\theoremstyle{remark}

\usepackage{overpic,wrapfig}
\usepackage[misc]{ifsym}
\usepackage{pifont}
\usepackage{natbib}

\makeatletter
\DeclareRobustCommand\onedot{\futurelet\@let@token\@onedot}
\def\@onedot{\ifx\@let@token.\else.\null\fi\xspace}

\makeatother

\newcommand{\cmark}{\ding{51}}%
\usepackage{enumitem}

\crefname{algorithm}{Alg.}{Algs.}
\Crefname{algocf}{Algorithm}{Algorithms}
\crefname{section}{Sec.}{Secs.}
\Crefname{section}{Section}{Sections}
\crefname{table}{Tab.}{Tabs.}
\Crefname{table}{Tab.}{Tabs.}
\crefname{figure}{Fig.}{Figs.}
\Crefname{figure}{Fig.}{Figs.}
\crefname{equation}{Eq.}{Eqs.}
\Crefname{equation}{Equation}{Equations}
\crefname{appendix}{Appx.}{Appxs.}
\Crefname{appendix}{Appendix}{Appendices}

%% file: 0.abstract.tex
Reward design is of great importance for solving complex tasks with reinforcement learning. Recent studies have explored using image-text similarity produced by vision-language models (VLMs) to augment rewards of a task with visual feedback. A common practice linearly adds VLM scores to task or success rewards without explicit shaping, potentially altering the optimal policy. Moreover, such approaches, often relying on single static images, struggle with tasks whose desired behavior involves complex, dynamic motions spanning multiple visually different states. Furthermore, single viewpoints can occlude critical aspects of an agent's behavior. To address these issues, this paper presents Multi-View Video Reward Shaping (MVR), a framework that models the relevance of states regarding the target task using videos captured from multiple viewpoints. MVR leverages video-text similarity from a frozen pre-trained VLM to learn a state relevance function that mitigates the bias towards specific static poses inherent in image-based methods. Additionally, we introduce a state-dependent reward shaping formulation that integrates task-specific rewards and VLM-based guidance, automatically reducing the influence of VLM guidance once the desired motion pattern is achieved. We confirm the efficacy of the proposed framework with extensive experiments on challenging humanoid locomotion tasks from HumanoidBench and manipulation tasks from MetaWorld, verifying the design choices through ablation studies.

%% file: 1.introduction.tex
\section{Introduction}\label{sec:intro}
\begin{figure*}[t!]
  \centering
  \includegraphics[width=0.8\linewidth]{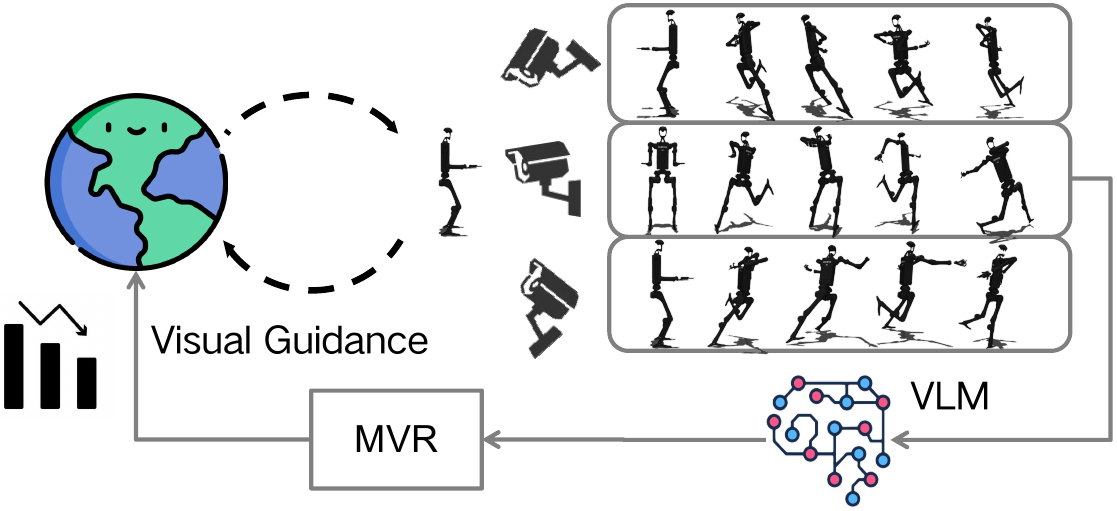}
  \caption{\textbf{The proposed MVR computes visual guidance using a VLM and videos collected from multiple viewpoints.}
  In this example, the task requires a humanoid robot to run forward.
  Being captured from different viewpoints, the image sequences encode complementary information and enable comprehensive evaluation of the agent's behaviors.
  This example also illustrates the pitfall of using image-text similarity for dynamic tasks—running requires rhythmic alternation of legs, but optimizing image-text similarity leads to realizing the pose that best matches ``running'' repeatedly.
  The shaping term prioritizes states that establish alternating leg cadence; once a stable gait and the target forward speed are achieved, its influence automatically decreases and the task reward takes precedence (see \cref{method}).}
\label{fig:teaser} % Keep label for \cref
\vskip -0.1in
\end{figure*}

% The current philosophy of reward design
When using reinforcement learning (RL), agents learn to optimize \emph{task rewards} encoding quantitative objectives.
For example, in the running task shown in \cref{fig:teaser}, the task rewards incentivize the tracking of a designated forward speed.
%Inspired by studies on animal locomotion~\cite{Hoyt1981,10.1242/jeb.01498}, task rewards are often combined with penalties for energy consumption to foster agile motion~\cite{pmlr-v164-fu22a,pmlr-v229-zhuang23a,10610200}.
In contrast, when learning new skills, humans can benefit from visual guidance tailored to the desired motion pattern, thus preventing suboptimal postures.
Inspired by this observation, we study the incorporation of visual feedback into task rewards in this paper.

With the development of \ac{VLM}s, an emerging paradigm is to leverage image-text similarity as rewards~\cite{baumli2023vision,rocamonde2024visionlanguagemodelszeroshotreward}, which guides agents to visit states that best match the textual description of a task.
However, this paradigm faces three shortcomings. Firstly, as single-image similarity cannot characterize dynamic motions, the generated rewards are mis-specified—they incentivize agents to visit the very states with maximal similarity score~\citep{fu2024furl}. As a result, agents cannot learn rhythmic movements involving repeated transitions between different states. Moreover, single viewpoints suffer from occlusion between robot limbs, creating view-dependent bias that further destabilizes dense guidance. Selecting the viewpoint with minimal occlusion often requires extensive human effort and domain knowledge. In addition, most methods linearly add single-view VLM scores to task or success rewards without explicit shaping, which can alter the optimal policy.

% Key idea of MVR
This paper presents \ac{MVR}, an online RL framework that overcomes these issues. As illustrated in \cref{fig:teaser}, MVR collects multi-view videos of agent behaviors and estimates state relevance from video-text similarity. This learned relevance constructs a \emph{state-dependent reward shaping} term that provides dense guidance early on and automatically diminishes as behavior improves, avoiding persistent conflict with task objectives.

We evaluate the effectiveness of MVR on a total of 19 tasks from HumanoidBench~\citep{sferrazza2024humanoidbench} and MetaWorld~\citep{yu2019meta}.
MVR consistently outperforms VLM-RM~\citep{rocamonde2024visionlanguagemodelszeroshotreward}, and RoboCLIP~\citep{sontakke2024roboclip}, which are existing VLM-based methods.
Furthermore, we validate the design of MVR in an ablation study and investigate the influence of using multiple views.
Lastly, we showcase the effectiveness of visual feedback for preventing suboptimal behaviors.
The contributions of this paper are as follows.
\begin{enumerate}
    \item We propose MVR, a framework for generating feedback from multi-view videos.
    \item We introduce a \emph{state-dependent reward shaping} formulation that integrates task rewards with VLM-based guidance and decays its influence as behaviour aligns with the target.
    \item We validate MVR’s effectiveness through simulated experiments on humanoid locomotion tasks in HumanoidBench and manipulation tasks in MetaWorld, and present a case study to showcase the importance of visual guidance.
\end{enumerate}

%% file: 2.related-work.tex
\section{Related work}
\textbf{VLM-based and Video-based Rewards\quad}
Recent work leverages VLMs to generate reward signals through image-text similarity~\citep{baumli2023vision, rocamonde2024visionlanguagemodelszeroshotreward, NEURIPS2023_aa933b5a,cui2024anyskill} or goal images~\citep{gao2023can}.
These methods typically treat the VLM as a frozen scoring function and then improve reward quality via VLM fine-tuning~\citep{fu2024furl}, binarization~\citep{huang2024dark}, or ranking-based objectives~\citep{wang2024rlvlmf}.

Beyond single-frame similarity, a growing body of work learns rewards directly from videos of task executions.
Rank2Reward~\citep{yang2024rank2reward} combines a temporal ranking-based reward model with an adversarial (GAIL-style) discriminator; PROGRESSOR~\citep{arxiv_2411.17764} learns a progress estimator and applies reward-weighted regression on real robots; and GVL~\citep{arxiv_2411.04549} uses a Gemini-based VLM to provide zero-shot progression rewards for downstream applications.
Generative video models such as Diffusion Reward~\citep{huang2024diffusionreward} and GenFlowRL~\citep{yu2025genflowrl} define rewards from conditional video diffusion or object-centric flow, while Video-Language Critic (VLC)~\citep{alakuijala2025vlc} fine-tunes a CLIP4Clip-style backbone so that a video-language critic can provide transferable rewards for training language-conditioned RL policies.
REDS~\citep{kim2025reds} learns subtask-aware dense rewards from segmented demonstrations, and ReWiND~\citep{zhang2025rewind} trains a video-language transformer to produce dense language-conditioned rewards and demonstrates real-world RL fine-tuning.

However, most of these approaches define rewards directly on raw image or video observations, without learning an explicit state-space relevance function, or assume access to curated demonstration datasets.
Classical example-guided controllers such as DeepMimic~\citep{peng2018deepmimic} and recent state-only imitation--emulation frameworks such as CIMER~\citep{han2024cimer} also rely on expert trajectories rather than re-using unlabeled online rollouts.
RoboCLIP~\citep{sontakke2024roboclip} is closer to our setting and uses video-text similarity but provides only sparse trajectory-level rewards.
MVR differs by learning state relevance from multi-view videos through paired comparisons and introducing state-dependent shaping that automatically decays, providing dense guidance during online RL while remaining compatible with task rewards.

\textbf{RL with Foundation Models\quad}
Foundation models have been extensively integrated into RL pipelines~\citep{xu2024survey,moroncelli2024integrating,kawaharazuka2024real}.
VLMs enable planning~\citep{pan2024vlp, patel2023pretrained}, success detection~\citep{pmlr-v232-du23b}, and representation learning~\citep{chen2024vision}, though most rely on single-frame inputs unsuitable for dynamic tasks.
LLMs facilitate planning~\citep{ichter2022do, shinn2024reflexion}, exploration~\citep{du2023guiding}, code-based policy generation~\citep{liang2023code}, and reward design~\citep{ma2024eureka, xie2024textreward}.
World models provide alternative approaches for reward generation~\citep{escontrela2023video} and model-based RL~\citep{wu2024ivideogpt}.
Our work specifically addresses the unique challenges of using pre-trained VLMs for continuous motion tasks, proposing practical solutions for video-based multi-view reward learning.
% NEURIPS2023_7ce1cbed,seo2022reinforcement

%Most of these approaches rely on a single modality, either image or language. In this paper, we focus on using video-language-aligned VLMs to learn reward functions, effectively leveraging multimodal information.

% Our work falls within the broader category of leveraging foundation models in RL, which is an active field with many exciting advances~\citep{ma2024eureka,xie2024textreward}. Apart from VLM, other forms of foundation models such as large language models (LLM) have also been used in many different ways, including planning~\citep{ichter2022do}, task decomposing with grounding~\citep{ichter2022do}, generating code as policy/skill~\citep{liang2023code,szot2024large}, reward design ~\citep{ma2024eureka,xie2024textreward} etc.. In this paper, we focus on a complementary perspective by highlighting the potential issues of using a pre-trained VLM in RL and proposing practical remedies.

%% file: 3.problem_statement.tex
\section{Problem Statement}\label{problem_statement}
\textbf{Markov Decision Process\quad}
The Markov Decision Process (MDP)~\cite{sutton2018reinforcement} is defined as $\langle\mathcal{S}, \mathcal{A}, P, r, \gamma \rangle$, where $\mathcal{S}$ and $\mathcal{A}$ denote the state space and the action space. 
$P:\mathcal{S}\times\mathcal{A}\to\Delta(\mathcal{S})$ is the transition probability of states, and $r:\mathcal{S}\times\mathcal{A}\to\mathbb{R}$ is the reward function.
%We use $r^\text{task}$ for the task rewards and $r^\text{VLM}$ for rewards  to denote the task rewards.
$\gamma\in(0,1)$ is the discount factor.
An MDP prescribes an interaction protocol in which an agent repeatedly receives states from an environment, samples actions from its policy $\pi(a|s):\mathcal{S}\to\Delta(A)$, and receives rewards.
The value function $v^\pi(s)=\mathbb{E}\left[\sum_{t=1}^\infty \gamma^{t-1}r_t\right]$ characterizes the expected discounted return of $\pi$ starting from state $s$.
The objective of RL is to find the optimal policy $\pi^*$ such that $v^{\pi^*}(s)\geq v^\pi(s)$ for any state $s$.

% \begin{figure}[t]
%   \centering
%   \includegraphics[width=\linewidth]{figures/loss.pdf}
%     \caption{\textbf{The two objectives for learning state relevance.}  $f^\text{MVR}$ consist of a state encoder $g^\text{state}$ and a vector $g^\text{rel}$.
%     The objective $L_\text{matching}$ aims to match the paired comparisons between state sequences computed with $f^\text{MVR}$ to the paired comparisons induced by video-text similarity scores. 
%     The objective $L_\text{reg}$ aligns the similarity structure between state sequences to the similarity structure between videos.
%     This dual approach—matching paired comparisons and regularizing state representations—ensures that the MVR framework effectively learns state relevance while mitigating the impact of visual discrepancies and viewpoint changes.}
%     \label{fig:loss}
% \vskip -0.1in
% \end{figure}

\textbf{The Learning Problem\quad}
The video-text similarity computed by models such as ViCLIP~\citep{wang2024internvid} is denoted by $\psi^\text{VLM}$ and characterizes how well a video matches a text description.
During policy learning, we periodically sample state sequences $\mathbf{s}$, render them into videos $\mathbf{o}$ from different viewpoints, and compute the video-text similarity scores $\psi^\text{VLM}(\mathbf{o},\ell)$, where $\ell$ is the task description.
For notational convenience, we also use $\psi^\text{VLM}(\mathbf{o},\mathbf{o}')$ for the similarity between two videos $\mathbf{o}$ and $\mathbf{o}'$.
These steps result in a dataset $\mathcal{D}=\{(\mathbf{s}, \mathbf{o}, \psi^\text{VLM}(\mathbf{o},\ell))\}$.
Throughout the paper, we use bold symbols (e.g., $\mathbf{s}$) for sequences and plain symbols (e.g., $s$) for single timesteps.
%This dataset is periodically updated with new states generated during the learning process.
Given $\mathcal{D}$, we aim to learn a \emph{state relevance model} $f^\text{MVR}:\mathcal{S}\to\mathbb{R}$ with parameters $\theta^\text{MVR}$ to model the relevance of states for the task of interest.
With $f^\text{MVR}$, we derive a reward function $r^\text{VLM}:\mathcal{S}\to\mathbb{R}$ as visual feedback for the agent, complementing the task rewards $r^\text{task}$. 

%% file: 4.method.tex
\section{Multi-view Video Reward Shaping}\label{method}

The proposed MVR is illustrated in \cref{fig:framework}.
\Cref{subsec:relevance_learning} explains how the relevance model $f^\text{MVR}$ is learned, and \cref{subsec:policy_learning} introduces the proposed reward function and the entire framework.

\subsection{Learning State Relevance from Multi-view Videos}\label{subsec:relevance_learning}

\textbf{Challenges\quad} Operating in state space $\mathcal{S}$ enables efficient reward computation without per-step video rendering, which is crucial for million-step training.
However, this design raises two coupled challenges.
First, there is a semantic gap between states (proprioception, joint angles) and visual features (textures, colors), which prohibits directly regressing video-text similarity from states.
Second, MVR renders videos from one of four viewpoints ($0^\circ$, $90^\circ$, $180^\circ$, and $270^\circ$), which introduces systematic viewpoint biases: for instance, frontal views in \cref{fig:teaser} often yield higher similarities due to better visibility, creating spurious correlations unrelated to behavioral quality.
In the remainder of this subsection, we address the semantic gap with \emph{Matching Paired Comparisons} and the viewpoint-induced bias with \emph{Regularizing State Representations}.

\textbf{Matching Paired Comparisons\quad} To tackle the semantic gap between states and videos, we propose to match comparisons in the two domains instead of regressing raw similarity scores. 
Being trained with contrastive objectives, the video-text similarity scores of two videos reflect the probability that one video better matches the task than the other. 
\emph{We claim that such orderings should be preserved between the corresponding state sequences.}
Specifically, the probability that a video $\mathbf{o}$ better matches the task prompt $\ell$ than another video $\mathbf{o}'$ under the Bradley-Terry (BT) model \citep{10.2307/2334029} is given by 
\begin{equation}
    h_\text{vid}(\mathbf{o},\mathbf{o'}) = \sigma(\psi^\text{VLM}(\mathbf{o},\ell) - \psi^\text{VLM}(\mathbf{o}',\ell)),
\end{equation}
\noindent where $\sigma(x)=(1+\exp(-x))^{-1}$ is the sigmoid function.
Similarly, the probability that a state sequence $\mathbf{s}$ is better than another state sequence $\mathbf{s}'$ is
\begin{equation}
    h_\text{state}(\mathbf{s}, \mathbf{s}')=\sigma\Bigl(\frac{1}{n(\mathbf{s})}\sum_{s\in\mathbf{s}}f^\text{MVR}(s)-\frac{1}{n(\mathbf{s'})}\sum_{s\in\mathbf{s'}}f^\text{MVR}(s)\Bigr),
\end{equation}
\noindent where $n(\mathbf{s})$ is the length of $\mathbf{s}$.
Given two samples $(\mathbf{s}, \mathbf{o}, \psi^\text{VLM}(\mathbf{o}, \ell))$ and $(\mathbf{s'}, \mathbf{o'}, \psi^\text{VLM}(\mathbf{o'}, \ell))$, to preserve the ordering between videos and state sequences, we propose to minimize: 
\begin{equation}
L_\text{matching} = -h_\text{vid}(\mathbf{o}, \mathbf{o'})\log(h_\text{state}(\mathbf{s},\mathbf{s'})) - \bigl(1- h_\text{vid}(\mathbf{o}, \mathbf{o'})\bigr)\log\bigl(1-h_\text{state}(\mathbf{s},\mathbf{s'})\bigr).
\label{eq:matching}
\end{equation}
By minimizing this cross-entropy, $f^\text{MVR}$ can capture the relevance of states for the task without being distracted by the visual details.
It is akin to the objective of preference-based RL~\cite{NIPS2017_d5e2c0ad}, but with the crucial difference that we fit $f^\text{MVR}$ to probabilities rather than binary labels.
Importantly, $(\mathbf{o}, \mathbf{o'})$ can originate from different viewpoints because each video shares the same underlying state sequence; sampling pairs across views enlarges the comparison dataset without requiring additional rendering.

\textbf{Regularizing State Representations\quad}
The second challenge is the viewpoint-induced bias in $\psi^\text{VLM}$: different camera angles can change similarity scores for the same underlying behavior.
A straightforward solution is to randomize the choice of viewpoint, but relevance learning then becomes difficult as we use more viewpoints, since we will have fewer samples from any single informative view.
\emph{To tackle this data scarcity while reducing viewpoint bias, we enhance relevance learning by regularizing state representations with the similarity structure between video embeddings}.
Specifically, we parameterize $f^\text{MVR}$ as $f^\text{MVR}(s) = \langle g^\text{rel}, g^\text{state}(s)\rangle$, where $g^\text{state}:\mathcal{S}\to\mathbb{R}^{d}$ is a state encoder and $g^\text{rel}\in\mathbb{R}^d$ is a learnable vector. The scalar $d$ is the dimension of the state embeddings.
This decomposition decouples representation learning from relevance scoring: $L_\text{reg}$ shapes a shared state representation, while $L_\text{matching}$ selects a single relevance direction $g^\text{rel}$ in that space, rather than entangling multi-view alignment and task-specific ranking in a single monolithic scalar predictor.
Then, we use the following regularization term for relevance learning:

%minimize the difference between the similarity of state sequences and video sequences:
\begin{equation}
    L_\text{reg} = \left|\psi^\text{VLM}(\mathbf{o}_i, \mathbf{o}_j) -  \langle \bar{g}^\text{state}(\mathbf{s}_i), \bar{g}^\text{state}(\mathbf{s}_j)\rangle\right|,
    \label{eq:reg}
\end{equation}
\noindent where $\bar{g}^\text{state}(\mathbf{s})=n(\mathbf{s})^{-1}\sum_{s\in\mathbf{s}}g^\text{state}(s)$ is the representation of a state sequence.
This regularization leads to averaged state representations from which view-specific interference is reduced. 

This regularization produces viewpoint-invariant state representations by aligning state and video similarity structures.
While $L_\text{matching}$ preserves orderings, $L_\text{reg}$ anchors the learned embeddings, together stabilizing multi-view learning.

Finally, the overall objective for relevance learning is: 
\begin{equation}
    L_\text{rel} = L_\text{matching} + L_\text{reg}.
    \label{eq:total}
\end{equation}
% mention the figure
% discuss the two loss
% mention the final loss for relevance learning

%\paragraph{Remarks} 
% Dicussion
% Difference with PbRL: alignment two modalities
% So we use the similarity socres, rather than hard labels
% While \cref{obj:bt} resembles the objective of preference-based RL~\cite{NIPS2017_d5e2c0ad},
% Difference with RL-VLM-F
%\Cref{framework} shows MVR's entire online learning framework.

\subsection{Policy Learning with \texorpdfstring{$r^\text{VLM}$}{rVLM}}\label{subsec:policy_learning}
\textbf{Desiderata of $r^\text{VLM}$\quad}
Similar to the case of image-text similarity, directly using state relevance as $r^\text{VLM}$ is prone to mis-specification.
Instead, we opt for matching the expectation of state relevance–$r^\text{VLM}$ should lead to matching the learner's state distribution with videos of the highest video-text similarity, thus leading to the desired behaviors.
Meanwhile, we argue that $r^\text{task}$ should gradually dominate $r^\text{VLM}$ as the learned behaviors resemble the desired one, which is crucial for avoiding the expensive VLM fine-tuning, especially when the VLM is not aligned with the target task.

\begin{figure*}[t!]
  \centering
  \includegraphics[width=\textwidth]{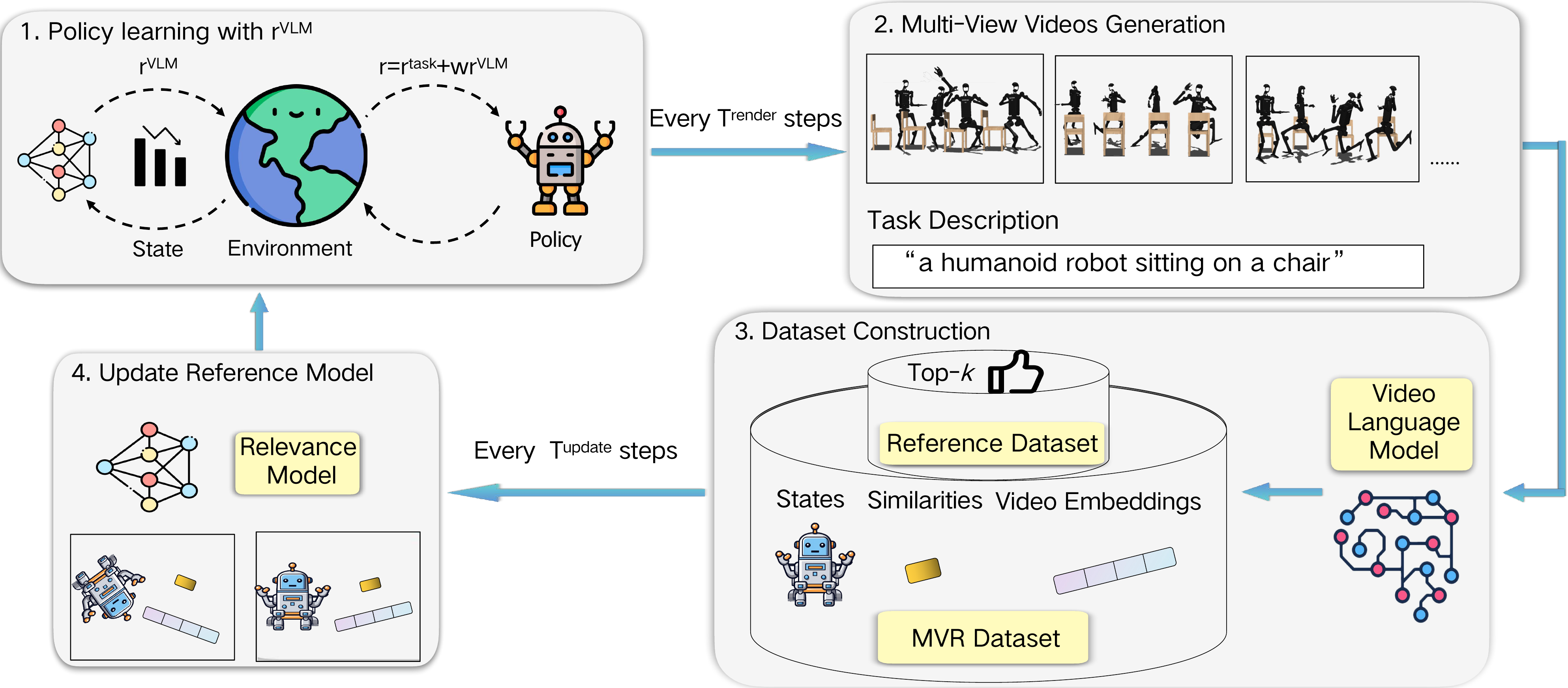}
    \caption{
    \textbf{The entire framework of the proposed MVR.}
    MVR periodically samples state sequences and renders them into videos from different viewpoints (step 2).
    It then queries a VLM for the similarity scores and video embeddings of the videos and augments its dataset $\mathcal{D}$ with state sequences, video embeddings, and similarity scores (step 3).
    Additionally, it keeps the state sequences with top-$k$ similarity scores in a reference set $\mathcal{D}^\text{ref}$. With $\mathcal{D}$, MVR updates the state relevance model (step 4). Lastly, using the latest state relevance model and the reference set, MVR computes visual feedback $r^\text{VLM}$ for the online RL agent (step 1), which is combined with task rewards $r^\text{task}$ through state-dependent reward shaping that automatically decays as the agent's behavior aligns with the reference set.}
    \label{fig:framework}
\vskip -0.1in
\end{figure*}

\textbf{The Proposed Reward Function\quad}
We first extend the notion of state relevance to policies.

\begin{definition}
Denote by $d^\pi(s)=(1-\gamma)\sum_{t=0}^{\infty} \gamma^t \, P(s_t{=}s\mid\pi)$ the state-occupancy distribution of policy $\pi$. The \emph{policy relevance} $h^\pi$ is the expectation of state relevance under $d^\pi$:
\begin{equation*}
    h^\pi = \sum_{s\in\mathcal{S}} f^\text{MVR}(s)\, d^\pi(s).
\end{equation*}
\end{definition}

Policy relevance characterizes the extent a policy matches the task.
Denote by $\pi^\ell$ the policy that best matches the task prompt $\ell$.
\emph{Our idea is to learn policies that maximize $r^\text{task}$ while being indiscernible from $\pi^\ell$}:
\begin{equation}
\max_\pi v^\pi + w\log (\sigma(h^\pi - h^{\pi^\ell})),
\label{eq:policy_obj}
\end{equation}
where $\log (\sigma(h^\pi - h^{\pi^\ell}))$ is the log-likelihood of $\pi$ being better than $\pi^\ell$ under the BT model.
$w\in\mathbb{R}$ is a weighting term.
%The function $\log(\sigma(x))$ is monotonically decreasing when $x \leq 0$, ensuring that the influence of $f^\text{MVR}$ weakens as $\pi$ becomes closer to $\pi^\ell$.
Using Jensen’s inequality, this log-likelihood can be lower-bounded as:
%noting that $h^\pi - h^{\pi^\ell} = \mathbb{E}_{s \sim d^\pi, s' \sim d^{\pi^\ell}}[f^\text{MVR}(s) - f^\text{MVR}(s')]$ and that $\log(\sigma(x))$ is concave, we can lower bound the log-likelihood term:
\begin{equation}\label{jesens}
\log \sigma(h^\pi - h^{\pi^\ell}) \geq \mathbb{E}_{s \sim d^\pi, s' \sim d^{\pi^\ell}} \left[ \log (\sigma(f^\text{MVR}(s) - f^\text{MVR}(s'))) \right].
\end{equation}

Thus, we can maximize \cref{eq:policy_obj} by using the shaped reward $r^\text{MVR}$ defined as
\begin{equation}
    r^\text{MVR}(s) \triangleq r^\text{task}(s) + w r^\text{VLM}(s),
\end{equation}
where $r^\text{VLM}$ is given by:
\begin{equation}
    r^\text{VLM}(s) \triangleq \mathbb{E}_{s'\sim\pi^\ell}\left[\log(\sigma(f^\text{MVR}(s)-f^\text{MVR}(s')))\right].
    \label{eq:r_vlm}
\end{equation}
% Discuss the tradeoff

This additive form serves as a \emph{state-dependent reward shaping} term rather than a fixed-weight sum of unrelated signals.
Since we do not have access to samples of $\pi^\ell$, \cref{eq:r_vlm} is computed using samples from a reference set $\mathcal{D}^\text{ref}$ that contains state sequences with top-$k$ video-text similarity scores (aggregated across viewpoints) collected during policy learning.
In literature, a similar problem is solved by learning a separate policy solely from VLM-based rewards~\citep{fu2024furl}, yet such approach fails when $r^\text{VLM}$ alone cannot solve the task.
Moreover, sampling from $\mathcal{D}^\text{ref}$ bears an interesting analogy to human skill learning: trying to repeat our previous good trials is an effective learning strategy for consolidating our understanding of a skill.

\paragraph{Automatic shaping.} As the policy improves, $r^\text{MVR}$ naturally relinquishes control to $r^\text{task}$: when the state distribution induced by $\pi$ aligns with the reference set $\mathcal{D}^\text{ref}$, the expectation in \cref{eq:r_vlm} approaches zero, causing $r^\text{VLM}$ to vanish automatically. This provides strong early guidance that naturally decays as behaviors improve, avoiding persistent conflicts with task objectives.

% To address this, we argue that \emph{$r^\text{VLM}$ should encourage the agent to visit relevant states but gradually diminish as the agent approaches those states}.
% This entails a flexible tradeoff between $r^\text{task}$ and $r^\text{VLM}$—while both reward types influence the agent's behavior in non-relevant states, the influence of both types of rewards exist for non-relevant states, $r^\text{task}$ will dominate when the agent’s motion aligns closely with the desired goal.
% Drawing analogy with human skill learning, this means to elicit the correct motion pattern with $r^\text{VLM}$ but switch to quantitative objective once the pattern is acquired.

\textbf{The Entire Framework\quad} \Cref{fig:framework} illustrates MVR's operation.
During policy learning, MVR renders the agent's latest trajectories into videos at a frequency $T^\text{render}$.
It then queries a VLM to obtain video-text similarity scores and video embeddings, augmenting the dataset $\mathcal{D}$ (lines 10–13).
$\mathcal{D}^\text{ref}$ is updated at the same time.
At a frequency $T^\text{update}$, MVR updates $f^\text{MVR}$ using \cref{eq:total} and samples in $\mathcal{D}$ (line 16–18).
Importantly, for off-policy algorithms such as TQC~\citep{kuznetsov2020controlling}, the rewards in the agent’s replay buffer must be recalculated after $f^\text{MVR}$ is updated.
The full pseudocode of MVR is provided in Algorithm~\ref{algo} in the appendix.

% \begin{figure*}[ht!]
%   \centering
%   \includegraphics[width=\linewidth]{tasks.pdf}
%   \caption{\textbf{The tasks considered in our experiments.}
%   These tasks requires a Unitree robot to perform dynamic motion (run, walk, stair, and slide) or static pose (stand, balance\_simple, sit\_simple, balance\_hard, and sit\_hard) given limited observation.}
%   \label{fig:tasks}
% \end{figure*}

%% file: 5.experiments.tex
\section{Experiments}\label{sec:experiments}
This section presents the task performance of MVR, followed by an ablation study for its components and results for the influence of using multiple views.
We also showcase the importance of combining visual guidance with task rewards using a case study.
\subsection{Setup}
% Setting and Tasks
\textbf{Tasks\quad}
We evaluate MVR on a total of 19 tasks from two robotics benchmarks: HumanoidBench~\citep{sferrazza2024humanoidbench} and MetaWorld~\citep{yu2019meta}.
Tasks from HumanoidBench are well-suited for exploring the use of VLMs in skill learning, as they encompass static posture generation (stand, balance\_simple, sit\_simple, balance\_hard, and sit\_hard) and dynamic motion generation (run, walk, stair, and slide).
% The tasks vary in difficulty. The walk, stair, and slide tasks have the same observations but differ in terrain.
Meanwhile, tasks from MetaWorld (hammer, push-wall, faucet-close, push-back, stick-pull, handle-press-side, push, shelf-place, window-close, peg-unplug-side) are used to evaluate performance in fine-grained manipulation tasks across diverse visual contexts. 
See \cref{HumanoidBench,MetaWorld} for detailed descriptions of the tasks considered in this paper. 
Among the 27 tasks available in HumanoidBench, we follow the benchmark authors and prior work by focusing on the nine locomotion and posture tasks listed above, excluding loco-manipulation tasks and the most challenging locomotion task \emph{Hurdle}, which do not yield meaningful learning progress within a 10M-step budget.
For MetaWorld, we adopt the CW10 subset from Continual World~\citep{wolczyk2021continual}, which contains ten single-object manipulation tasks that are commonly evaluated under a 1M-step budget.
\textbf{Evaluation Metrics\quad}
For HumanoidBench tasks, agents are trained for ten million environment steps, and we report the mean and standard deviation of the episodic returns over three random seeds. A higher score indicates better performance.
For MetaWorld tasks, agents are trained for one million environment steps, and we report the mean and standard deviation of the task success rates over five random seeds.
In our ablation studies on HumanoidBench, to showcase the relative effect of MVR's components, we normalize the performance of MVR's variants using the performance of MVR.
In addition, we report the average rank of methods to assess the overall performance for a domain. The lower the better.

\textbf{Alternative Methods\quad}
For methods only using $r^\text{task}$, we select two recent RL algorithms: TQC~\citep{kuznetsov2020controlling} and DreamerV3~\citep{hafner2023dreamerv3}.
For existing methods that leverage VLMs, we report results for VLM-RM~\citep{rocamonde2024visionlanguagemodelszeroshotreward} and RoboCLIP~\citep{sontakke2024roboclip}.
VLM-RM combines $r^\text{task}$ with image-text similarity scores\footnote{While \citet{rocamonde2024visionlanguagemodelszeroshotreward} only uses VLM-based rewards for pose generation, it is difficult to solve locomotion tasks without $r^\text{task}$.}.
RoboCLIP uses video-text similarity but generates a single reward value for each trajectory.
Since VLM-RM and RoboCLIP use different RL algorithms in their original implementation, we implement both on top of TQC to ensure fair comparison with MVR.
All VLM-based baselines, including RoboCLIP, shape their policies with both the task reward and their respective VLM-derived signals so that differences stem purely from the multimodal guidance design.
%In particular, both MVR and VLM-RM are built upon TQC, enabling us to evaluate: (1) the effectiveness of VLMs in robot skill learning and (2) the limitations of methods relying solely on image-text similarity. 

\begin{table*}[t!]
\centering
\caption{\textbf{\ac{MVR} outperforms VLM-RM and RoboCLIP for HumanoidBench tasks.} 
The first four tasks are dynamic tasks and the last five are static tasks.
We compare MVR with TQC~\citep{kuznetsov2020controlling}, DreamerV3~\citep{hafner2023dreamerv3}, VLM-RM~\citep{rocamonde2024visionlanguagemodelszeroshotreward}, and RoboCLIP~\citep{sontakke2024roboclip}. The DreamerV3 originate from the TD-MPC2 benchmark suite~\citep{hansen2024tdmpc2}.
MVR performs the best for five tasks and has the best rank averaged over all tasks.
VLM-RM performs the best for two tasks.
RoboCLIP cannot perform well for these tasks.\protect\footnotemark}
\label{Performance}
\vskip 0.1in
\resizebox{\textwidth}{!}{%
\begin{tabular}{lcccccc} 
\toprule
Task & \ac{MVR} & TQC & VLM-RM  & RoboCLIP & DreamerV3 & Success / Max Return \\
\midrule
 Walk    & $\mathbf{927.47\pm1.83}$~\cmark & $510.58 \pm 299.16$ & $535.35 \pm 355.18$ & $737.34 \pm 194.62$~\cmark & $800.2 \pm 158.7$~\cmark & 700 / 1000 \\ 
  Run     & $\mathbf{749.23\pm56.82}$~\cmark & $647.87\pm186.98$ & $14.93 \pm 1.11$ & $501.15 \pm 179.71$ & $633.8 \pm 222.4$ & 700 / 1000 \\
  Stair   & $208.60\pm166.22$ & $\mathbf{282.95\pm120.54}$ & $44.96 \pm 4.12$ & $211.33 \pm 56.25$ & $131.1 \pm 43.6$ & 700 / 1000 \\ 
  Slide   & $\mathbf{735.03\pm142.85}$~\cmark & $514.91 \pm 106.36$ & $163.13 \pm 41.22$ & $494.20 \pm 21.66$ & $436.5 \pm 200.1$ & 700 / 1000 \\ 
\midrule
Stand   & $\mathbf{918.55\pm 29.30}$~\cmark & $576.59 \pm 371.0$ & $728.69 \pm 102.19$ & $849.73 \pm 108.86$~\cmark & $622.7 \pm 404.8$ & 800 / 1000 \\ 
  Sit\_Simple & $861.07\pm19.35$~\cmark & $822.07\pm101.28$~\cmark & $293.72 \pm 65.46$ & $553.90 \pm 244.51$ & $\mathbf{891.4 \pm 38.4}$~\cmark & 750 / 1000 \\ 
  Sit\_Hard & $\mathbf{756.67\pm108.79}$~\cmark & $511.85\pm155.45$ & $322.95 \pm 45.70$ & $559.38 \pm 192.00$ & $433.4 \pm 355.9$ & 750 / 1000 \\ 
  Balance\_Simple & $299.87\pm58.45$ & $256.79\pm54.23$ & $\mathbf{654.42 \pm 335.19}$ & $215.81 \pm 18.19$ & $19.8 \pm 7.0$ & 800 / 1000 \\ 
  Balance\_Hard & $95.78\pm11.03$ & $92.35\pm10.76$ & $\mathbf{107.36 \pm 24.93}$ & $97.03 \pm 18.13$ & $45.9 \pm 27.4$ & 800 / 1000 \\ 
\midrule
Average Rank & \textbf{1.67} & 3.11 & 3.78 & 2.89 & 3.56 & \textemdash \\
\bottomrule
\end{tabular}%
}
\caption*{\scriptsize \cmark~indicates that the mean return exceeds the official HumanoidBench success threshold.}
\end{table*}
\footnotetext{Success thresholds follow the official HumanoidBench specification~\citep{sferrazza2024humanoidbench}; the maximum returns correspond to the theoretical upper bounds defined in the benchmark.}
% Add a new table for Metaworld results
\begin{table}[h]
\scriptsize
\centering
\caption{\textbf{MVR achieves higher success rate on Metaworld tasks.}
We use the results for DreamerV3 reported by TD-MPC2 benchmark suite~\citep{hansen2024tdmpc2}.
The three methods that incorporate visual guidance, MVR, VLM-RM, and RoboCLIP, all outperform TQC and DreamerV3.
Within them, MVR achieves the highest success rate, which is followed by RoboCLIP.
These results support our claim for the importance of using visual feedback and derive the feedback from multi-view videos.}
\label{tab:metaworld_results}
\resizebox{0.8\textwidth}{!}{%
\begin{tabular}{lccccc}
\toprule
Task & MVR & TQC & VLM-RM & RoboCLIP & DreamerV3 \\
\midrule
hammer & $0.20 \pm 0.24$ & $0.20 \pm 0.40$ & $\mathbf{0.40 \pm 0.49}$ & $0.00 \pm 0.00$ & $0.10 \pm 0.17$ \\
push-wall & $\mathbf{0.60 \pm 0.49}$ & $0.10 \pm 0.20$ & $0.40 \pm 0.37$ & $\mathbf{0.60 \pm 0.37}$ & $0.00 \pm 0.00$ \\
faucet-close & $\mathbf{1.00 \pm 0.00}$ & $\mathbf{1.00 \pm 0.00}$ & $\mathbf{1.00 \pm 0.00}$ & $\mathbf{1.00 \pm 0.00}$ & $0.13 \pm 0.15$ \\
push-back & $0.50 \pm 0.45$ & $\mathbf{0.60 \pm 0.37}$ & $0.30 \pm 0.40$ & $0.20 \pm 0.40$ & $0.37 \pm 0.38$ \\
stick-pull & $\mathbf{0.20 \pm 0.24}$ & $0.10 \pm 0.20$ & $0.00 \pm 0.00$ & $0.00 \pm 0.00$ & $0.00 \pm 0.00$ \\
handle-press-side & $\mathbf{1.00 \pm 0.00}$ & $\mathbf{1.00 \pm 0.00}$ & $\mathbf{1.00 \pm 0.00}$ & $\mathbf{1.00 \pm 0.00}$ & $0.77 \pm 0.17$ \\
push & $0.10 \pm 0.20$ & $0.10 \pm 0.20$ & $0.00 \pm 0.00$ & $\mathbf{0.30 \pm 0.40}$ & $0.00 \pm 0.00$ \\
shelf-place & $\mathbf{0.20 \pm 0.24}$ & $0.00 \pm 0.00$ & $0.00 \pm 0.00$ & $\mathbf{0.20 \pm 0.25}$ & $0.00 \pm 0.00$ \\
window-close & $\mathbf{1.00 \pm 0.00}$ & $\mathbf{1.00 \pm 0.00}$ & $\mathbf{1.00 \pm 0.00}$ & $\mathbf{1.00 \pm 0.00}$ & $0.61 \pm 0.16$ \\
peg-unplug-side & $0.60 \pm 0.37$ & $0.30 \pm 0.40$ & $0.60 \pm 0.49$ & $\mathbf{0.80 \pm 0.40}$ & $0.77 \pm 0.19$ \\
\midrule
Average Rank & \textbf{1.50} & 2.20 & 2.40 & 2.00 & 3.90 \\
\bottomrule
\end{tabular}%
}
\vskip -0.1in
\end{table}

\begin{figure*}[t]
    \centering
    % 第一个子图
    \begin{subfigure}[t]{0.49\textwidth}
        \centering
        \includegraphics[width=\textwidth]{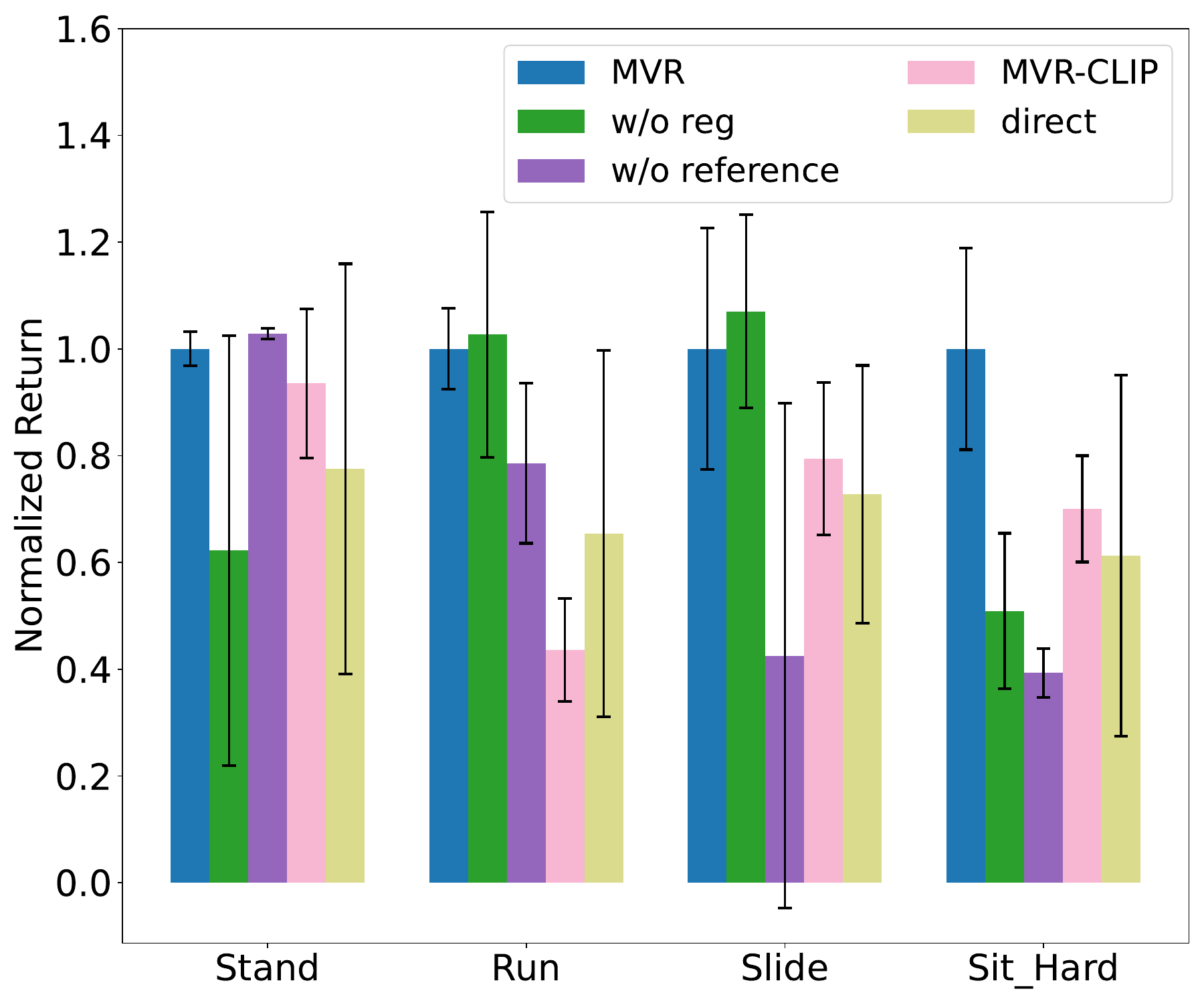}
    \caption{\textbf{The use of videos, \cref{eq:matching}, \cref{eq:reg}, and \cref{eq:r_vlm} are all useful.}
    The variant \emph{w/o reg} and \emph{w/o reference} stand for disabling \cref{eq:reg} and \cref{eq:r_vlm}. \emph{MVR-CLIP} and \emph{direct} means using images and fitting the $f^\text{MVR}$ to the similarity scores rather than using \cref{eq:matching}.
    No variant has consistently good performance as MVR does.}
    \label{fig:ablation}
    \end{subfigure}
    \hfill
    % 第二个子图
    \begin{subfigure}[t]{0.49\textwidth}
        \centering
        \includegraphics[width=\textwidth]{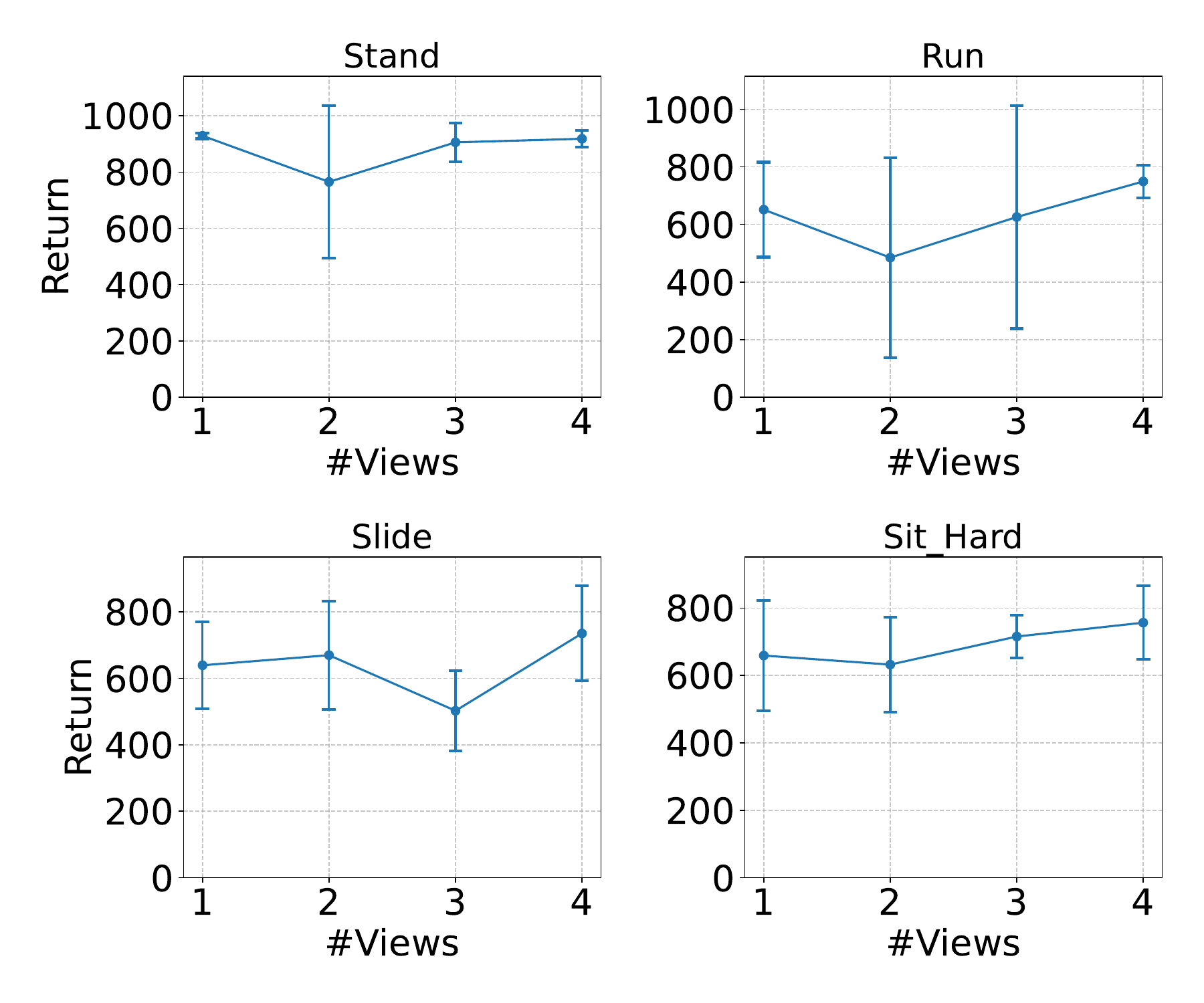}
\caption{\textbf{Using more views is beneficial.} These figures show the performance of MVR when using one to four views. For Stand, one view seems sufficient as the return is close to optimal. For other tasks, a general trend is that performance improves when using more views, though fluctuation exists.}
    \label{fig:views}
    \end{subfigure}
    \caption{Method ablation and the influence of the number of views.}
    \label{fig:all_ablation}
\vskip -0.1in
\end{figure*}

% \begin{figure}[h]
%   \centering
%   \includegraphics[width=\linewidth]{figures/ablation.pdf}
%     \caption{\textbf{The use of multi-view videos, the proposed learning objective \cref{eq:total}, and the proposed reward function \cref{eq:r_vlm} all contribute to the performance of MVR.}
%     We compare MVR with several variants that remove one of its component.
%     The performance of these variant is normalized by the performance of MVR for each task.
%     These results indicate that all the components are necessary for soliciting visual guidance.}
%     \label{fig:ablation}
% \vskip -0.1in
% \end{figure}

\textbf{Ablations\quad}
We compare MVR with several variants to validate its components.
The variant \textit{w/o reg} removes $L_\text{reg}$ from \cref{eq:total}.
For \textit{w/o reference}, we use the raw relevance $f^\text{MVR}(s)$ rather than \cref{eq:r_vlm}$\,$as visual guidance, ablating the state-dependent expectation matching (and its natural decay).
The variant \textit{MVR-CLIP} uses image-text similarity computed by a CLIP model.
For the variant \textit{direct}, we train $f^\text{VLM}$ by directly fitting it to the video-text similarity scores. 
Moreover, we compare the results using one to four views to analyze the influence of using multiple viewpoints.
Lastly, we compare the results of using ViCLIP-L (428M parameters) and ViCLIP-B (150M parameters) to study the influence of VLM size.
Additional ablations on reward components and alternative temporal pooling strategies are summarised in \cref{tab:rvlm_only,tab:attention_pooling}.

\textbf{Implementation Details\quad}
MVR renders one trajectory out of every nine into a video ($T^\text{render}=9$).
For each rendered episode we draw one viewpoint uniformly at random; across training this rotates through all cameras without exceeding the rendering/encoding budget of a single-view pipeline.
For each video, MVR extracts segments of length 64 ($T^\text{video}=64$) and queries the ViCLIP model~\citep{wang2024internvid} for video-text similarity scores.
The relevance model is updated every 100,000 environment steps ($T^\text{update}=100,000$) with early stopping.
The reference set $\mathcal{D}^\text{ref}$ keeps the top-$k$ most relevant state sequences observed so far with $k=10$ for all experiments, i.e., only the highest-quality trajectories remain in memory.
The hyperparameter $w$ in \cref{eq:policy_obj} is selected from $\{0.01, 0.1, 0.5\}$ via grid-search.
In contrast, VLM-RM queries VLMs at each policy learning step, which is not scalable for the tasks considered here. 
Therefore, we adopt the same rendering frequency as MVR and fit a reward model for policy learning.
Both MVR and RoboCLIP use ViCLIP-L (428M parameters). VLM-RM uses CLIP-ViT-H-14 (986M parameters). More details can be found in \cref{baselines}.

% \begin{figure}[t]
%   \centering
%   \includegraphics[width=0.8\linewidth]{figures/hyperablation.pdf}
%     \caption{\textbf{The best value of $w$ in \cref{eq:policy_obj} is task-dependent}.
%     A general trend is that smaller values are preferred when the agent needs to perform rapid motion.
%     }
%     \label{fig:hyperablation}
% \vskip -0.1in
% \end{figure}

\begin{table}[h]
\centering
\caption{\textbf{MVR can aggregate information from multiple views.} This table compares the task performance of using four different viewpoints only and random viewpoints. While the best viewpoint is task-dependent, using random viewpoints (MVR) has the best performance for Slide and Sit\_Hard and second-best performance for Run, indicating that MVR can aggregate multi-view information.}
\label{tab:multiview_comparison}
\resizebox{0.8\textwidth}{!}{%
\begin{tabular}{lccccc}
\toprule
Task & $0^\circ$ & $90^\circ$ & $180^\circ$ & $270^\circ$ & MVR \\
\midrule
Stand   & $914.17 \pm 32.54$  & $929.12 \pm 10.58$  & $912.20 \pm 26.66$  & $\mathbf{931.10 \pm 32.08}$  & $918.55 \pm 29.30$  \\
Run    & $611.43 \pm 294.44$ & $651.23 \pm 164.96$ & $\mathbf{875.33 \pm 6.44}$  & $567.91 \pm 159.47$ & $749.23 \pm 56.82$  \\
Slide   & $559.59 \pm 0.00$   & $639.28 \pm 131.48$ & $647.18 \pm 154.73$ & $657.64 \pm 205.81$ & $\mathbf{735.03 \pm 142.85}$\\
Sit\_Hard & $528.14 \pm 176.60$ & $659.10 \pm 163.77$ & $452.15 \pm 210.89$ & $723.72 \pm 112.11$ & $\mathbf{756.67 \pm 108.79}$ \\
\bottomrule
\end{tabular}%
}
\vskip -0.1in
\end{table}

\begin{figure*}[t]
  \centering
  \includegraphics[width=\textwidth]{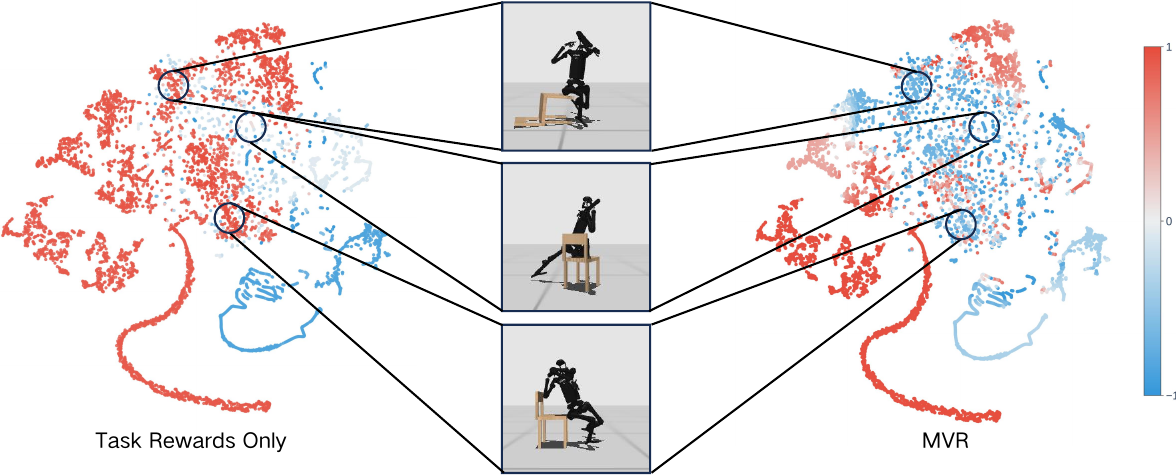}
    \caption{\textbf{MVR identifies suboptimal states through reward shaping.}
    We visualize states generated by a TQC agent for the Sit\_Hard task and annotate them with $r^\text{task}$ (left) and the shaped reward $r^\text{MVR} = r^\text{task} + w r^\text{VLM}$ (right) computed by a trained MVR agent.
    While task rewards are high when the agent is close to the chair, MVR's visual guidance component assigns low values to improper sitting poses (sitting on chair's leg, leaning, or sitting at the edge), effectively shaping the reward landscape to discourage these visually suboptimal but task-rewarding states.
    }
    \label{fig:demo}
\vskip -0.1in
\end{figure*}

\subsection{Results}\label{subsec:task_performance}
\textbf{Proficiency for Soliciting Visual Guidance\quad}
\Cref{Performance} presents results for the task performance of HumanoidBench tasks.
MVR has the best performance for five tasks.
VLM-RM outperforms other methods for two tasks, but RoboCLIP cannot outperform others for any tasks.
Notably, MVR has the best average rank for these tasks, corroborating its proficiency for soliciting visual guidance.
The comparison among MVR, VLM-RM, and TQC reveals an interesting insight.
VLM-RM performs worse than TQC for three dynamic tasks but surpasses it for three static tasks.
In contrast, MVR outperforms TQC for three dynamic tasks and all static tasks.
Thus, generating visual guidance from videos not only applies to challenging dynamic tasks but is also more effective for static tasks.

~\cref{tab:metaworld_results} shows the results for MetaWorld tasks. MVR attains the highest average success rate and has the best average rank, followed by RoboCLIP. Despite RoboCLIP's strong overall performance (second in average success rate and rank), it scored zero on two specific tasks, indicating potential instability. VLM-RM has a higher average success rate than TQC but a worse average rank, indicating inferior performance on certain tasks.

The performance margins between multi-view methods are smaller on MetaWorld than on HumanoidBench for two reasons. First, MetaWorld episodes are short (500 steps) and focus on single tabletop objects, where sparse trajectory-level feedback already provides useful exploration hints. Second, the long-horizon HumanoidBench locomotion suite demands dense state-conditioned guidance across multiple, occasionally occluded joints, which explains the larger performance gaps observed in \cref{Performance}.

% MVR achieves a higher average success rate (0.54) compared to the TQC baseline (0.44) over five random seeds, demonstrating the applicability of our approach to a different robot (Sawyer arm) and task domain (manipulation).
% Specifically, MVR shows notable improvements in tasks such as \emph{push-wall} (0.60 vs 0.10), \emph{shelf-place} (0.20 vs 0.00), and \emph{peg-unplug-side} (0.60 vs 0.30). Performance is comparable on several other tasks (e.g., \emph{hammer}, \emph{faucet-close}, \emph{handle-press-side}, \emph{push}, \emph{window-close}).
% On \emph{push-back}, TQC performed slightly better, indicating that the visual reward from MVR was not always beneficial or might have required further tuning of the weight $w$ for that specific task.
% Overall, the results suggest that the visual feedback provided by MVR, learned through multi-view videos, can aid in learning fine-grained manipulation skills. The improvements in tasks requiring precise interaction, like \emph{shelf-place} and \emph{peg-unplug-side}, suggest that MVR can help the agent learn subtle visual cues relevant to successful task completion.

Meanwhile, readers may notice that, among the 19 tasks considered in this paper, MVR performs worse than TQC for four tasks, indicating that using $r^\text{MVR}$ might compromise the optimization of $r^\text{task}$. A possible reason is that the underlying VLM is trained on mostly human activities rather than robotics data, and we leave the few-shot extension of MVR for future work.
\textbf{Model Ablation\quad} We now discuss the results presented in \cref{fig:ablation} from three perspectives.
Firstly, the inferior performance of \emph{direct} and the unstable performance of \emph{w/o reg} indicate that both \cref{eq:matching} and \cref{eq:reg} are essential for learning the relevance model.
Moreover, the decreased performance of MVR-CLIP shows that assessing agent behaviors with videos is critical for deriving visual guidance, further confirming the insight discussed above.
Lastly, the variant \emph{w/o reference} is outperformed by MVR for three tasks, confirming the efficacy of our proposed reward function \cref{eq:r_vlm}.
Altogether, these results demonstrate the effectiveness of components in MVR.

\begin{wraptable}{l}{0.45\textwidth}
\vspace{-1.5em}
\scriptsize
\centering
\caption{\textbf{MVR with small and large VLMs.}}
\label{tab:vlm_comparison}
\begin{tabular}{lcc}
\toprule
Task & ViCLIP-L (428M) & ViCLIP-B (150M)\\
\midrule
Stand    & $\mathbf{918.55} \pm \mathbf{29.30}$  & $917.97 \pm \mathbf{30.71}$  \\
Run      & $\mathbf{749.23} \pm \mathbf{56.82}$  & $517.25 \pm \mathbf{383.12}$ \\
Slide    & $735.03 \pm \mathbf{142.85}$ & $\mathbf{778.84} \pm \mathbf{24.80}$  \\
Sit\_Hard & $\mathbf{756.67} \pm \mathbf{108.79}$ & $689.50 \pm \mathbf{234.37}$ \\
\bottomrule
\end{tabular}
\vspace{-1.5em}
\end{wraptable}
\noindent
As shown in Table \ref{tab:vlm_comparison}, a comparison between the ViCLIP-B and ViCLIP-L models reveals that the larger ViCLIP-L model yields better performance in three of the four evaluated tasks, confirming the common hypothesis of the scaling law.
In addition, we present the results for using different RL algorithms in MVR in \cref{subsec:algorithms} and the results for using different values for the weight $w$ of $r^\text{VLM}$ in \cref{subsec:weight}.

\textbf{The Influence of Views\quad} \Cref{fig:views} presents the results of MVR when using different numbers of views.
While using one view is enough for the Stand task, it is beneficial to increase the number of views for other tasks, which confirms our claim for using multi-view videos for computing $r^\text{VLM}$.
Meanwhile, \cref{tab:multiview_comparison} shows the results of different views and compares them with MVR, which selects views uniformly at random.
The best choice of view is task-dependent, which can be explained by the difference in the required motion patterns.
For example, for Sit\_Hard, the two side views $90^\circ$ and $270^\circ$ are more effective than the rest two because it is easier to determine the robot-chair distance from side views.
Notably, moving from one view to four orthogonal views raises the mean returns on Run, Slide, and Sit\_Hard by +98, +96, and +98 respectively, pushing them past the HumanoidBench success thresholds (700, 700, 750) and turning previously failing policies into successful ones. Since the HumanoidBench returns are capped at 1000, these gains amount to roughly 10\% of the total attainable return.
Interestingly, we can obtain competitive or even better results by randomly sampling viewpoints, which highlights MVR's capability in aggregating information in multi-view videos.

\textbf{Case Study\quad} Finally, we showcase the significance of visual guidance using the Sit\_Hard.
\Cref{fig:demo} is a t-SNE~\citep{JMLR:v9:vandermaaten08a} visualization of 20 episodes produced by a trained TQC agent.
We annotate the states with $r^\text{task}$ and $r^\text{MVR}$ computed by a trained MVR agent in \cref{fig:demo}.
The rewards are normalized to the range [-1, 1] for visualization purpose.

A general trend in \cref{fig:demo} is that $r^\text{task}$ and $r^\text{MVR}$ align in many parts of the state space.
For example, both reward functions assign high rewards to the bottom left part of the space and low rewards to the bottom right.
However, upon closer inspection of the regions where the two reward functions diverge, we gain valuable insights.
Specifically, the center region corresponds to undesirable sitting poses, such as sitting on the chair's leg (the top example), leaning on the chair (the middle example), and sitting at the chair's edge (the bottom example).
Although these poses receive high values of $r^\text{task}$ due to the minimal distance between the robot and the chair, they are inherently unstable and suboptimal.
Notably, MVR is able to identify them, which explains its superior performance over TQC in this task.
Without visual guidance, a much larger number of online samples will be required for recognizing such \emph{visually} suboptimal behaviors by their long-term consequences.
This emphasizes the importance of visual guidance in learning general and versatile skills.

\textit{Quantitative analyses.} \Cref{subsec:reward_alignment_app} details the correlation between the learned shaping signals and binary success, and also documents the temporal decay of $r^\text{VLM}$ that drives the automatic handover to $r^\text{task}$.

%% file: 6.conclusions.tex
\section{Conclusions}
\label{sec:conclusion}
In this paper, we present a novel framework named Multi-view Video Reward to generate visual guidance using VLMs for RL agents. 
In contrast to existing attempts that use VLMs to generate rewards, MVR leverages videos collected from multiple viewpoints to evaluate agent behaviors, which enables it to characterize dynamic motions and overcome occlusions.
Moreover, we propose a state-dependent reward shaping formulation that integrates task rewards with VLM-based guidance and automatically reduces the influence of the auxiliary signal as the behavior aligns with the desired motion pattern.
We evaluate MVR for nine humanoid tasks and ten manipulation tasks. 
MVR has the best average rank when compared to existing methods that use VLMs and methods that only use task rewards, confirming its superiority for soliciting visual guidance for RL agents.
Furthermore, we confirm the design of MVR with an ablation study and investigate the influence of using multiple views.
Lastly, we present a case study to showcase the importance of visual feedback for fostering the desired motion of a task.

\paragraph{Ethics Statement.} This work uses only simulated benchmarks, does not collect human data, and does not interact with sensitive or safety-critical systems. To the best of our knowledge, it poses no foreseeable ethical, privacy, or security risks.

\paragraph{Reproducibility Statement.} \Cref{subsec:relevance_learning,algo} specify the training objectives, optimization steps, and update schedule for MVR. Implementation details covering architectures, environment settings, rendering schedules, and all hyperparameters appear in \Cref{subsec:implementation_details}, while additional ablations and alternative backbones are summarized in \Cref{subsec:additional_ablations,subsec:algorithms}. \Cref{sec:additional_results} compiles the evaluation tables cited in the main text. The submission bundles the Hydra configuration files, evaluation scripts, and fixed random seeds required to reproduce every experiment.

%% file: appendix.tex
\renewcommand\thefigure{A\arabic{figure}}
\setcounter{figure}{0}
\renewcommand\thetable{A\arabic{table}}
\setcounter{table}{0}
\renewcommand\theequation{A\arabic{equation}}
\renewcommand\thealgorithm{A\arabic{algorithm}}
\setcounter{equation}{0}
\pagenumbering{arabic}% resets `page` counter to 1
\renewcommand*{\thepage}{A\arabic{page}}
\setcounter{footnote}{0}
\resetlinenumber[1]

\appendix
\onecolumn

\section{Experimental Setup}
\label{Experimental Setting}

\subsection{Implementation Details}
\label{subsec:implementation_details}

In this study, we adopted the JAX implementation of TQC from the SBX library~\citep{raffin2021stable} as our baseline RL algorithm. The parameters of this algorithm were subsequently fine-tuned to address the complexities inherent in humanoid robot control tasks. Our proposed Multi-view Video Rewards (MVR) framework builds upon this TQC foundation. The experiments were conducted using the most recent version of the HumanoidBench simulation environment~\citep{sferrazza2024humanoidbench}.

All computational experiments were performed on NVIDIA RTX 4090 GPUs. Under this hardware configuration, each MetaWorld task required approximately 25 minutes to complete, while each HumanoidBench task took roughly 4 hours. The specific versions of key software packages utilized throughout our experiments are enumerated below:

\begin{itemize}
    \item Python 3.11
    \item Torch 2.3.1
    \item TorchVision 0.18.1
    \item JAX 0.4.37
    \item JAXLib 0.4.36
    \item Flax 0.10.2
    \item Gymnasium 0.29.1
    \item SBX-RL 0.18.0
\end{itemize}

\subsection{MVR Algorithm Pseudocode}
\label{appx:pseudocode}

For completeness, we present the full pseudocode of the proposed MVR framework below; see \cref{fig:framework} in the main text for the high-level pipeline.

\begin{algorithm}[t!]
\caption{The Pseudocode of MVR}
\label{algo}
\resizebox{0.8\linewidth}{!}{%
\begin{minipage}{\linewidth}
\begin{algorithmic}[1]
\STATE \textbf{Input:} textual goal description $\ell$, environment steps $T^\text{env}$, video sampling frequency $T^\text{render}$, model update frequency $T^\text{update}$
    \STATE Initialize the policy $\pi$ and relevance model $f^\text{MVR}$
    \STATE Initialize the replay buffer of MVR: $\mathcal{D}=\emptyset$.
    \STATE Initialize the reference set: $\mathcal{D}^\text{ref}=\emptyset$.
\FOR{t = 0, 2, ...,  $T^\text{env}$}
    \STATE Observe state $s_t$, apply action $a_t\sim\pi(\cdot|s_t)$ and observe the next state $s_{t+1}$ and reward $r^\text{task}_t$.
    \STATE Compute $r^\text{VLM}_t$ and form the shaped reward $r^\text{MVR}_t = r^\text{task}_t + w r^\text{VLM}_t$.
    \STATE Use $s_t$, $a_t$, $s_{t+1}$, and $r^\text{MVR}_t$ to update $\pi$.
    \IF{t > 0}
	        \IF{t \% $ T^\text{render}$ = 0}
	            \STATE Render the last episode into videos.
	            \STATE Augment $\mathcal{D}$ with the latest state sequences, videos, and their video-text similarity scores: $\mathcal{D}= \mathcal{D}\cup\{(\mathbf{s}, \mathbf{o},\psi^\text{VLM}(\mathbf{o}, \ell))\}$.
	            \STATE Switch to the next viewpoint.
	            \STATE Update the reference set $\mathcal{D}^\text{ref}$ if necessary.
	        \ENDIF
        \IF{t \% $ T^\text{update}$ = 0}
        \STATE Use data in $\mathcal{D}$ to update $f^\text{MVR}$ with \cref{eq:total}.
        \ENDIF
    \ENDIF
\ENDFOR
\end{algorithmic}
\end{minipage}}
\end{algorithm}

\subsection{HumanoidBench}
\label{HumanoidBench}

Humanoid robots offer significant promise for assisting humans in a multitude of environments and tasks, primarily due to their human-like morphology, which grants them flexibility and adaptability. HumanoidBench~\citep{sferrazza2024humanoidbench} is a significant contribution in this area, offering a simulated benchmark with 27 distinct whole-body control tasks. These tasks are designed to present unique and substantial challenges, including those requiring intricate long-horizon control and sophisticated coordination. As demonstrated by~\citep{sferrazza2024humanoidbench}, even state-of-the-art RL algorithms face difficulties with many of these tasks. For our study, we excluded tasks that were identified as poorly designed (e.g., the 'powerlift' task lacked appropriate reward signals for lifting the barbell), resulting in a set of nine valid tasks for evaluation.

A brief description of each selected HumanoidBench task is provided below.

\paragraph{Stand} The robot's objective is to maintain an upright standing posture for the duration of the trial. This requires dynamic adjustments to ensure stability, compensating for environmental changes or internal balance shifts. This task evaluates the robot's capacity for stable positioning, a fundamental skill for stationary operations and transitions between movements.

\paragraph{Walk} This task involves the robot maintaining a consistent forward velocity of approximately $1 \text{ m/s}$ along the global x-axis, without falling. It challenges the robot's balance, coordination, and its ability to generate efficient forward locomotion, crucial for navigation in varied settings.

\paragraph{Run} Here, the robot must achieve and maintain a forward running speed of $5 \text{ m/s}$. This higher-speed task demands advanced coordination, strength, and precise control over leg movements. It involves not just maintaining velocity but also managing the complex dynamics of transitioning between walking and running gaits.

\paragraph{Stair} The robot is tasked with traversing a repeating sequence of upward and downward stairs at a steady speed of $1 \text{ m/s}$. This requires the robot to adapt its gait and step height for each stair while preserving balance and forward momentum, simulating real-world obstacle navigation.

\paragraph{Slide} In this task, the robot navigates an alternating sequence of upward and downward inclines (slides) at a speed of $1 \text{ m/s}$. The primary challenge lies in adapting to varying surface inclinations and maintaining stability on potentially slippery or uneven terrain, testing real-time gait adjustment capabilities.

\paragraph{Sit\_Simple} The robot must sit on a chair positioned closely behind it. This task tests the robot's environmental perception, motion planning for the sitting sequence, and controlled execution of the sit-down action while maintaining stability.

\paragraph{Sit\_Hard} This task increases the complexity of sitting by randomizing the chair's initial position. Furthermore, the chair can move due to forces exerted by the robot, meaning its position and orientation can change dynamically. The robot must continuously adapt its approach, handling the uncertainty of the chair's movement while maintaining balance.

\paragraph{Balance\_Simple} The robot is required to maintain balance on an unstable board that pivots on a fixed spherical point. The core challenge is to remain stable on a surface that can tilt in any direction, demanding continuous postural adjustments and center of mass control.

\paragraph{Balance\_Hard} This task extends the balancing challenge by introducing a moving pivot point for the unstable board. This adds a layer of complexity, as the robot must simultaneously adjust for the board's instability and the dynamic movement of its pivot, necessitating a high degree of coordination and real-time responsiveness.

Prompts derived from these task descriptions are summarized in \cref{tab:robot_prompts}.

\begin{table}[ht!]
\centering
\caption{Text instructions for humanoid robot tasks.}
\label{tab:robot_prompts}
\begin{tabular}{ll}
  \toprule % Changed from \hline for booktabs style
  \textbf{Prompt} & \textbf{Description} \\
  \midrule % Changed from \hline for booktabs style
  Stand & a humanoid robot standing \\
  Walk & a humanoid robot walking \\
  Run & a humanoid robot running \\
  Stair & a humanoid robot walking \\
  Slide & a humanoid robot walking \\
  Sit\_Simple & a humanoid robot sitting on a chair \\
  Sit\_Hard & a humanoid robot sitting on a chair \\
  Balance\_Simple & a humanoid robot balancing on the board \\
  Balance\_Hard & a humanoid robot balancing on the board \\
  \bottomrule % Changed from \hline for booktabs style
\end{tabular}
\end{table}

\subsection{MetaWorld}
\label{MetaWorld}

Meta-World is an open-source simulated benchmark for meta-RL and multi-task learning consisting of 50 distinct robotic manipulation tasks. We selected 10 Metaworld tasks in CW10~\citep{wolczyk2021continual} because their difficulty is suitable for solving in 1M steps.
A brief introduction to each selected task is summarized below.

\paragraph{Hammer} In this task, the robot gripper is required to pick up a hammer and use it to strike a nail or a similar target on a surface. This task tests the robot's abilities in grasping, tool manipulation, force control, and aiming accuracy.
\paragraph{Push-Wall} This task requires the robot gripper to push an object, such as a puck or cylinder, towards a target location. However, a wall obstructs a direct path, so the robot must navigate the object around or over the wall. This task challenges path planning, obstacle avoidance, and pushing mechanics.
\paragraph{Faucet-Close} The robot gripper must turn a faucet handle clockwise to close it. This involves precise grasping of the handle and applying the correct rotational force. It tests fine motor control and the understanding of object affordances.
\paragraph{Push-Back} In this task, the robot gripper is required to push a wooden block backward to a designated goal position. This primarily tests the robot's ability to apply controlled force in a specific direction and manage the displacement of the object.
\paragraph{Stick-Pull} The robot gripper needs to grasp a stick and pull it towards itself or out of a fixture. This task assesses grasping strength, pulling force control, and potentially the ability to dislodge an object from its holder.
\paragraph{Handle-Press-Side} This task requires the robot gripper to press a handle downwards, approaching it from the side. This tests the robot's ability to apply force at a specific angle and direction, often simulating actions like opening a door or activating a mechanism.
\paragraph{Push} This is a fundamental pushing task where the robot gripper must push an object, typically a cylinder or puck, to a target location on a flat surface. It evaluates the robot's basic pushing capabilities and goal-oriented movement.
\paragraph{Shelf-Place} The robot gripper is tasked with picking up an object, such as a cube, and then accurately placing it onto a designated spot on a shelf. This involves a sequence of grasping, lifting, 3D path planning, and precise placement.
\paragraph{Window-Close} In this task, the robot gripper needs to close a window. This might involve pushing a sliding window or manipulating a latch or handle, depending on the window type. The task tests interaction with larger articulated objects and achieving a specific state change in the environment.
\paragraph{Peg-Unplug-Side} The robot gripper must grasp a peg that is inserted into a hole and then unplug it by pulling it out sideways. This requires precise grasping, applying force in a specific lateral direction, and successfully disengaging the parts.

Prompts derived from these task descriptions are summarized in Table \ref{tab:robot_prompts_metaworld}.

\begin{table}[ht!]
\centering
\caption{Text instructions for MetaWorld robot tasks.}
\label{tab:robot_prompts_metaworld}
\begin{tabular}{ll}
\toprule
\textbf{Prompt} & \textbf{Description} \\
\midrule
hammer & a robot gripper hammering a screw on the wall \\
push-wall & a robot gripper bypassing a wall and pushing a cylinder \\
faucet-close & a robot gripper rotating the faucet clockwise \\
push-back & a robot gripper pushing the wooden block backward \\
stick-pull & a robot gripper pulling a stick \\
handle-press-side & a robot gripper pressing a handle down sideways \\
push & a robot gripper pushing a cylinder \\
shelf-place & a robot gripper picking and placing a cube onto a shelf \\
window-close & a robot gripper pushing and closing a window \\
peg-unplug-side & a robot gripper unplugging a peg sideways \\
\bottomrule
\end{tabular}
\end{table}

\subsection{Architecture and Hyperparameters}
\label{Architecture and Hyperparameters}

\paragraph{RL Training}
Each humanoid task was trained for 10 million steps, and each metaworld task was trained for 1 million, following the setup outlined in \citep{sferrazza2024humanoidbench,hansen2024tdmpc2}. Samples were collected from 8 parallel environments, with the learning rate decaying from $6 \times 10^{-4}$ at the start to $5 \times 10^{-5}$ at the final step, ensuring training stability. The policy was updated by sampling from the replay buffer with a batch size of 256. Every 16 steps, MVR performs 16 gradient updates. Both actor and critic networks consist of three-layer neural networks, each with a width of 256 and a quantile count of 50, while the number of top quantiles to drop per network is set to 5. SDE~\citep{raffin2022smooth} was enabled to enhance the algorithm's exploration. The key parameters are summarized in \cref{tab:vlm_tqc_params}.

\paragraph{Compute cost}
For each rendered trajectory we sample only one viewpoint, so the wall-clock cost matches that of a single-view implementation. A HumanoidBench run takes roughly four hours on one NVIDIA RTX~4090 using MuJoCo's EGL backend (about 20~GB GPU memory). When GPU rendering is unavailable we fall back to the OSMesa backend, which requires approximately 2.5~GB of CPU memory. MetaWorld tasks are shorter; each run finishes in about 25 minutes and needs roughly 2~GB of GPU memory. These numbers are consistent with the settings reported in the official HumanoidBench repository.

\paragraph{Reward Model}
The proposed reward model is composed of two core components: a multi-layer backbone network and a parameterized predictor vector. As outlined in \cref{tab:vlm_tqc_params}, the backbone adopts a fully-connected architecture with two hidden layers of dimension 512, activated by ReLU nonlinearity. The predictor is implemented as a single learnable vector of dimension 512, initialized via Kaiming uniform distribution to stabilize gradient dynamics during training.
Formally, given an input state $\mathbf{s} \in \mathbb{R}^d$, the backbone first projects it into a latent embedding space through sequential transformations:
\begin{equation}
    \mathbf{h}_1 = \text{ReLU}(\mathbf{W}_1 \mathbf{s} + \mathbf{b}_1), \quad
    \mathbf{h}_2 = \mathbf{W}_2 \mathbf{h}_1 + \mathbf{b}_2
\end{equation}
where $\mathbf{W}_1 \in \mathbb{R}^{512 \times d}$, $\mathbf{W}_2 \in \mathbb{R}^{512 \times 512}$ are weight matrices. The final embedding $\mathbf{e} = \mathbf{h}_2 / \|\mathbf{h}_2\|_2$ is $\ell_2$-normalized along the feature dimension to constrain the output scale.
The predictor $\mathbf{p} \in \mathbb{R}^{512}$ operates as a direction-sensitive projection head. It is normalized as $\tilde{\mathbf{p}} = \mathbf{p} / \|\mathbf{p}\|_2$ and broadcasted across the batch dimension. The value $f^{\text{MVR}}$ is computed via cosine similarity between the normalized embedding and the predictor:
\begin{equation}
    f^{\text{MVR}} = \langle \mathbf{e}, \tilde{\mathbf{p}} \rangle = \sum_{i=1}^{512} e_i \cdot \tilde{p}_i
\end{equation}
This design bounds reward outputs within $[-1, 1]$, promoting training stability. Optionally, the model can return the intermediate embedding $\mathbf{e}$ for auxiliary tasks or interpretability analysis.

\begin{table*}[ht!]
\centering
\caption{Hyperparameters of MVR}
\label{tab:vlm_tqc_params}
\begin{tabular}{lcc}
\toprule
\textbf{Category} & \textbf{Hyperparameter} & \textbf{Value} \\
\midrule
\textbf{TQC Network Architecture} \\
& Policy Layers & [256, 256, 256] \\
& Critic Layers & [256, 256, 256] \\
& Quantile Count & 50 \\
\midrule
\textbf{Reward Model} \\
& Backbone & [512,512] \\
& Predictor & Parameter(512) \\
\midrule
\textbf{RL Training} \\
& Total Timesteps  $T^\text{env}_{humanoidbench}$ & $1 \times 10^7$ \\
& Total Timesteps $T^\text{env}_{metaworld}$ & $1 \times 10^6$ \\
& Number of Environments & 8 \\
& Learning Rate & lin\_$6 \times 10^{-4} \rightarrow 5 \times 10^{-5}$ \\ % Adjusted arrow
& Batch Size & 256 \\
& Buffer Size & $1 \times 10^6$\\
& Gradient Steps & 16 \\
& Tau $\tau$ & 0.01 \\
& Gamma $\gamma$ & 0.99 \\
& Sde Sample Frequency & 4 \\
& Top Quantiles to Drop Per Net & 5 \\
\midrule
\textbf{Reward Model Learning} \\
& Model & ViCLIP \\
& Image Resolution & $224 \times 224$ \\
& Reward Scale $w$ & $\{0.01,0.1,0.5\}_{\text{best}}$ \\ % Adjusted subscript
& Video Clip Length $T^\text{video}$ & 64 \\
& Relabel Interval & $1 \times 10^5$ \\
& Update Interval $T^\text{update}$ & $1 \times 10^5$ \\
& Video Sampling Frequency $T^\text{render}$ & 9 \\
& Encoding Batch Size & 32 \\
& Reward Dataset Size & $2 \times 10^4$ \\
\bottomrule
\end{tabular}
\end{table*}

\paragraph{Reward Model Learning}
MVR renders one out of every nine trajectories into videos with a resolution of $224 \times 224$. From each video, MVR extracts segments of length 64 and queries the ViCLIP model~\citep{wang2024internvid} for video-text similarity scores, using a batch size of 32. These scores are then added to the reward dataset, which has a size of $2 \times 10^4$. The relevance model is updated every 100{,}000 environment steps, with early stopping employed to prevent overfitting. The hyperparameter $w$ in \cref{eq:policy_obj} is selected from $\{0.01, 0.1, 0.5\}$ using grid search. MVR relabels the dataset every $1 \times 10^5$ steps. Key parameters are summarised in \cref{tab:vlm_tqc_params}.

\subsection{Top-/Bottom-Ranked Frames in Sit\_Hard}
\label{subsec:top_bottom_frames_app}

\begin{figure*}[t]
  \centering
  \includegraphics[width=\textwidth]{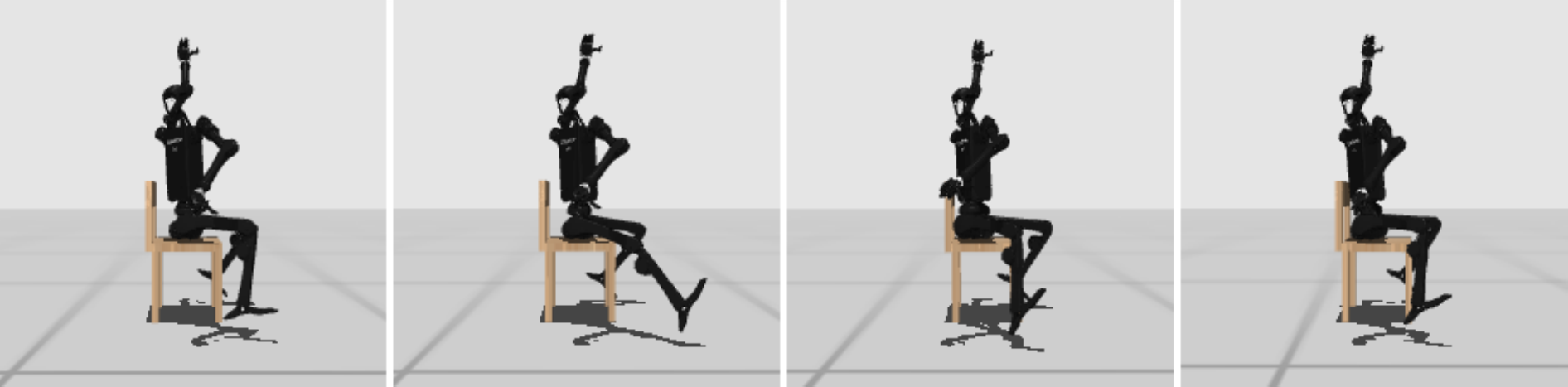}\\[2pt]
  \includegraphics[width=\textwidth]{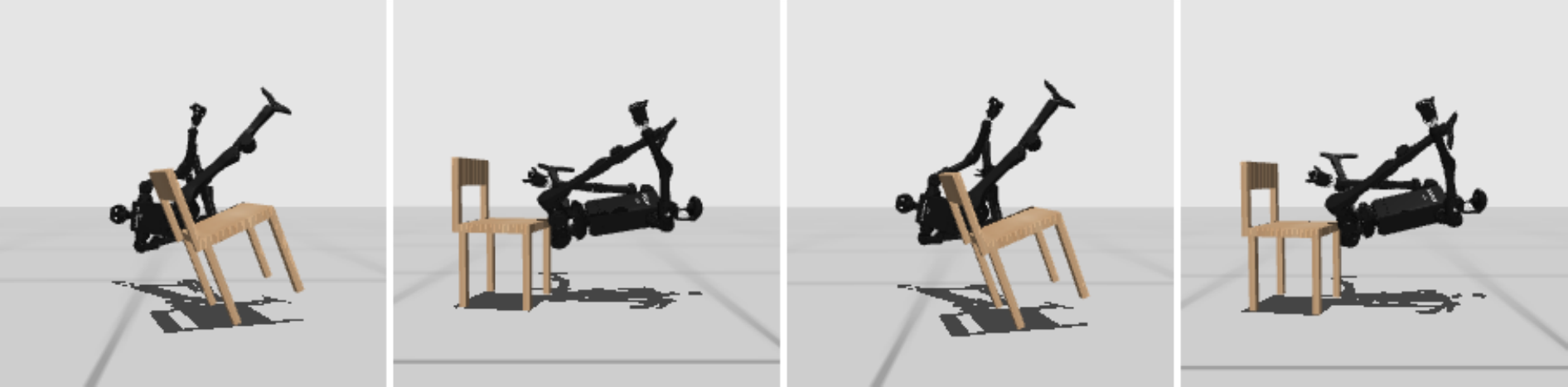}\\[6pt]
  \includegraphics[width=\textwidth]{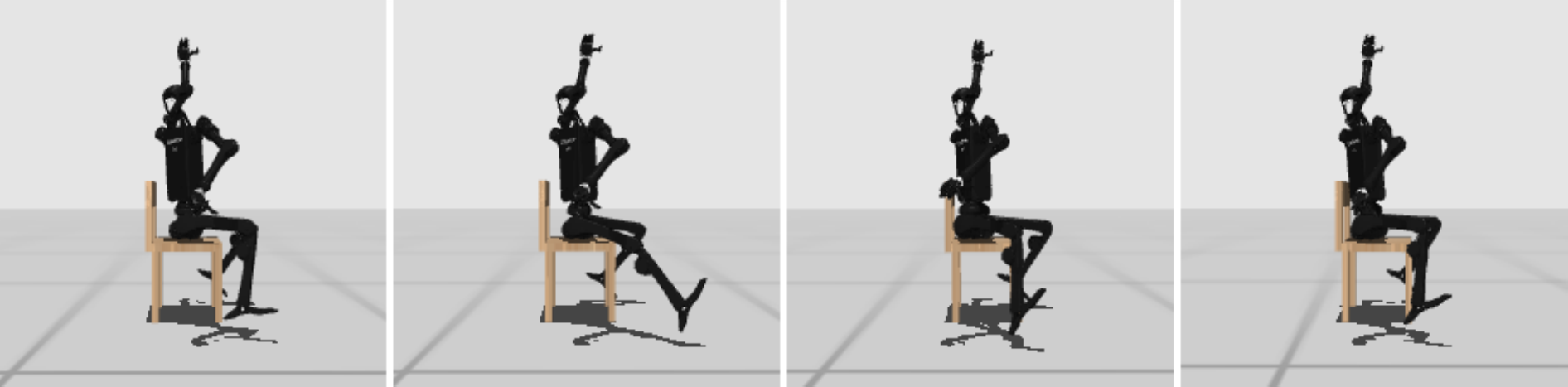}\\[2pt]
  \includegraphics[width=\textwidth]{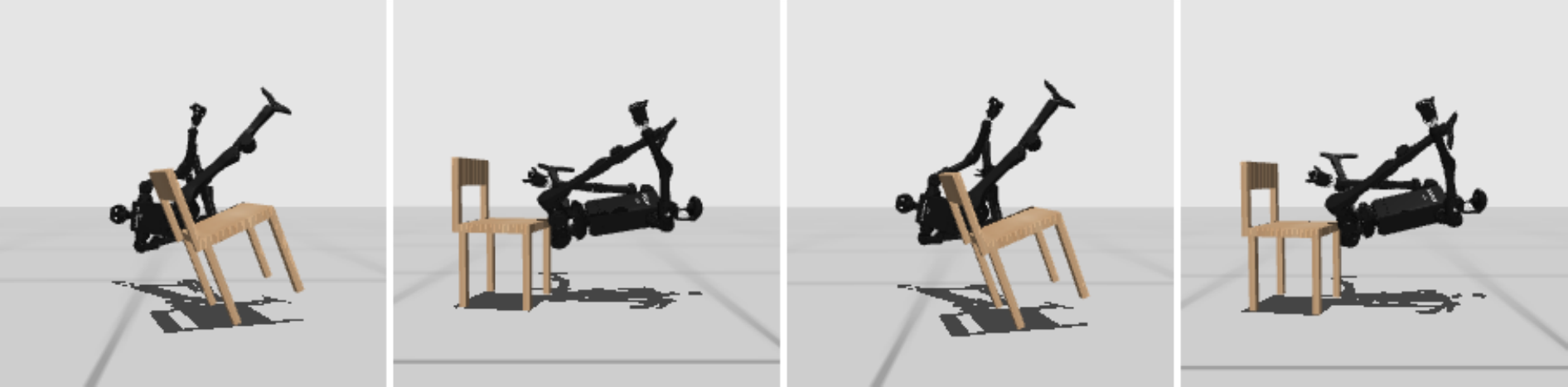}
  \caption{\textbf{Top-/bottom-ranked frames by $r^\text{task}$ and $r^\text{MVR}$ in the Sit\_Hard task.}
  The first two rows show the top and bottom frames under the task reward $r^\text{task}$, while the last two rows show the corresponding frames under the shaped reward $r^\text{MVR}$.
  For both rewards, the highest-ranked frames coincide with visually unambiguous successful sitting poses, confirming that MVR preserves the optimal behavior emphasized by the task reward.
  The bottom-ranked frames are dominated by clearly failed attempts under both signals, serving mainly as a sanity check that obviously bad states receive low values.
  Together with the qualitative analysis in \cref{fig:demo}, this suggests that the main effect of MVR is to reshuffle intermediate-reward states, down-weighting visually unstable but task-rewarding poses while leaving the optimal behavior unchanged.}
\label{fig:top_bottom_task_mvr}
\end{figure*}

To further clarify where MVR actively modifies the reward landscape, we visualize frames from the Sit\_Hard task ranked by the task reward $r^\text{task}$ and by the full shaped reward $r^\text{MVR} = r^\text{task} + w r^\text{VLM}$.
Figure~\ref{fig:top_bottom_task_mvr} shows the top- and bottom-ranked frames under both criteria.
The top-ranked frames for $r^\text{task}$ and $r^\text{MVR}$ consistently correspond to clearly successful behaviors in which the humanoid sits stably on the chair with an upright torso, indicating that the visual shaping term preserves the optimal behavior selected by the task reward.
The bottom-ranked frames for both rewards are dominated by catastrophic failures, where the robot has fallen off the chair or never approaches a seated configuration, confirming that both signals assign low value to clearly unsuccessful states even though the exact ordering among these failures is less critical for control.
This qualitative alignment on clearly successful behaviors is consistent with our goal that $r^\text{VLM}$ automatically decays once the policy reliably reaches the desired motion pattern.

In contrast, differences between $r^\text{task}$ and $r^\text{MVR}$ primarily emerge in the intermediate-reward regime.
As illustrated by the analysis in \cref{fig:demo}, clusters of states that receive relatively high $r^\text{task}$ but correspond to visually unstable sitting poses (e.g., sitting on a chair leg, slipping off the edge, or leaning heavily) are systematically down-weighted by $r^\text{MVR}$.
These mid-range corrections are precisely where MVR is active: the shaped reward preserves the ordering of clearly successful states while sharpening the ranking among borderline cases, assigning lower values to visually unstable yet task-rewarding poses and relatively higher values to visually robust configurations.

\subsection{VLM Baselines}
\label{baselines}

This section provides a detailed overview of the VLM baselines. \Cref{tab:models_used} summarizes the models used.\footnote{For fairness, we exclude methods such as RL-VLM-F~\citep{wang2024rlvlmf} that require repeated queries to large proprietary VLMs, which falls outside our automated large-scale evaluation setting.}

\begin{table*}[ht!]
\centering
\caption{Models Used in MVR, VLM-RM, and RoboCLIP}
\label{tab:models_used}
\begin{tabular}{l|l}
\toprule
\textbf{Algorithm} & \textbf{Model} \\
\midrule
\textbf{MVR} & ViCLIP \\
\midrule
\textbf{VLM-RM} & CLIP-ViT-H-14-laion2B-s32B-b79K \\
\midrule
\textbf{RoboCLIP} & ViCLIP \\
\bottomrule
\end{tabular}
\end{table*}

\paragraph{VLM-RM}
VLM-RM uses a pre-trained vision-language model (VLM) as a zero-shot reward model (RM) to represent tasks in natural language. The original implementation employs CLIP to train a MuJoCo humanoid robot for static tasks, such as kneeling, doing splits, and sitting cross-legged. In this paper, we use CLIP-ViT-H-14-laion2B-s32B-b79K to match the model size with ViCLIP. We select TQC instead of SAC because it demonstrates superior performance and serves as the RL backbone for MVR, eliminating performance discrepancies related to algorithm selection. We directly use MSE loss to fit the CLIP output, avoiding ranking loss to ensure maximum consistency with the original VLM-RM implementation. $r^\text{VLM-RM}$ is defined as follows:
\begin{equation}
    r^\text{VLM-RM}(s) \triangleq \mathbb{E}_{s'\sim\pi^\ell}\left[f^\text{VLM-RM}(s)\right].
    \label{eq:r_vlm_rm}
\end{equation}
where $f^\text{VLM-RM}(s)$ represents the reference function. The remaining parameters of VLM-RM align with MVR (see \cref{tab:vlm_tqc_params}).

\paragraph{RoboCLIP}
RoboCLIP generates rewards for online RL agents using pre-trained video-language models. It provides a sparse, end-of-trajectory reward based on the similarity between the agent's video and a target text description. The original version embedded inputs using S3D~\citep{xie2018rethinking} (pre-trained on HowTo100M~\citep{miech2019howto100m}) and used the unscaled scalar product of embeddings as the reward. In this paper, we adapt RoboCLIP by substituting ViCLIP for S3D and replacing the SAC algorithm with TQC to align with the MVR baseline for comparison. The algorithms and corresponding VLM models are summarized in \cref{tab:models_used}.

\section{Additional Results}
\label{sec:additional_results}

This section presents further experimental results to provide a more comprehensive evaluation of MVR and its components.

\subsection{Compatibility with Different RL Algorithms}
\label{subsec:algorithms}

To evaluate the general applicability of the MVR framework, we integrated it with a different RL algorithm, SAC, in addition to the TQC algorithm used in the main experiments. \Cref{tab:mvr_sac_tqc_comparison_booktabs} compares the performance of MVR-SAC against the baseline SAC algorithm, alongside the MVR-TQC and TQC results for reference. We further pair MVR with Simba~\citep{huang2024simba}, a large-scale off-policy learner, to confirm that the shaping term scales to stronger architectures.

\begin{table}[htbp]
\centering
\caption{\textbf{MVR yields greater performance improvements when integrated with the TQC algorithm compared to SAC on complex humanoid tasks.} This table presents average returns for MVR-SAC and MVR-TQC against their SAC and TQC baselines across selected HumanoidBench tasks.}
\label{tab:mvr_sac_tqc_comparison_booktabs}
\begin{tabular}{lcccc}
\toprule
\textbf{Task} & \textbf{MVR-SAC} & \textbf{SAC} & \textbf{MVR-TQC} & \textbf{TQC} \\
\midrule
Stand & $704.24 \pm 329.17$ & $581.77 \pm 371.68$ & $\mathbf{918.55 \pm 29.30}$  & $576.59 \pm 371.0$   \\
Run & $280.24 \pm 103.91$ & $299.39 \pm 314.65$ & $\mathbf{749.23 \pm 56.82}$  & $647.87 \pm 186.98$ \\
Slide & $219.79 \pm 176.13$ & $94.43 \pm 52.89$ & $\mathbf{735.03 \pm 142.85}$ & $514.91 \pm 106.36$ \\
Sit\_Hard & $172.46 \pm 124.35$ & $263.87 \pm 97.12$  & $\mathbf{756.67 \pm 108.79}$ & $511.85 \pm 155.45$ \\
\bottomrule
\end{tabular}
\end{table}

The results indicate that the performance benefit from VLM-based rewards provided by MVR is less pronounced when using SAC compared to TQC. By contrast, integrating MVR with Simba preserves the gains on the hardest locomotion tasks, demonstrating that multi-view shaping transfers to the latest high-capacity agents (\cref{tab:simba_followup}).

\begin{table}[h]
\scriptsize
\centering
\caption{\textbf{HumanoidBench returns for Simba with and without MVR shaping.} Simba numbers are taken from the original publication; Simba+MVR averages three seeds under the same horizon.}
\label{tab:simba_followup}
\begin{tabular}{lcc}
\toprule
Task & Simba & Simba+MVR \\
\midrule
Stand & $\mathbf{906.94}$ & $814.53$ \\
Run & $741.16$ & $\mathbf{817.69}$ \\
Slide & $577.87$ & $\mathbf{814.53}$ \\
Sit\_Hard & $783.95$ & $\mathbf{785.23}$ \\
\bottomrule
\end{tabular}
\end{table}

\subsection{VLM Backbone Ablation}
\label{subsec:vlm_backbone}

To assess the sensitivity of MVR to the choice of vision-language backbone, we additionally train an S3D-based variant of MVR on four HumanoidBench tasks and compare it with the ViCLIP-based version and the TQC baseline.
\Cref{tab:vlm_backbone} shows that MVR-S3D consistently outperforms TQC and achieves performance comparable to MVR-ViCLIP (slightly better on Run/Stand and slightly worse on Slide/Sit\_Hard), suggesting that our gains do not rely on a particular VLM.

\begin{table}[htbp]
\centering
\caption{\textbf{Effect of VLM backbone on MVR performance.} Returns on four HumanoidBench tasks for TQC, MVR with ViCLIP, and MVR with S3D.}
\label{tab:vlm_backbone}
\begin{tabular}{lccc}
\toprule
\textbf{Task} & \textbf{TQC} & \textbf{MVR-ViCLIP} & \textbf{MVR-S3D} \\
\midrule
Run      & $647.87 \pm 186.98$ & $749.23 \pm 56.82$  & $\mathbf{793.61 \pm 57.27}$ \\
Slide    & $514.91 \pm 106.36$ & $\mathbf{735.03 \pm 142.85}$ & $696.51 \pm 32.39$ \\
Stand    & $576.59 \pm 371.00$ & $918.55 \pm 29.30$  & $\mathbf{937.20 \pm 14.28}$ \\
Sit\_Hard & $511.85 \pm 155.45$ & $\mathbf{756.67 \pm 108.79}$ & $678.53 \pm 106.32$ \\
\bottomrule
\end{tabular}
\end{table}

\subsection{Occlusion-Heavy HumanoidBench Task}
\label{subsec:occlusion_pole}

To further assess MVR under severe occlusion, we evaluate the \emph{Pole} task from HumanoidBench, where the humanoid must move forward through a dense forest of thin poles without colliding with them.
In this scenario, different body parts are frequently occluded from any single viewpoint, making multi-view coverage especially important.
As shown in \Cref{tab:occlusion_pole}, MVR substantially outperforms the task-reward baseline TQC on this occlusion-heavy locomotion task.

\begin{table}[htbp]
\centering
\caption{\textbf{HumanoidBench \emph{Pole} task under heavy occlusion.} Returns for TQC and MVR.}
\label{tab:occlusion_pole}
\begin{tabular}{lcc}
\toprule
\textbf{Task} & \textbf{TQC} & \textbf{MVR} \\
\midrule
Pole & $601.62 \pm 192.34$ & $\mathbf{956.43 \pm 13.02}$ \\
\bottomrule
\end{tabular}
\end{table}

\subsection{Wrist-Mounted Camera Views in MetaWorld}
\label{subsec:metaworld_wrist_camera}

To complement the exocentric setup used in the main MetaWorld experiments, we consider a variant that incorporates a wrist-mounted hand camera when computing VLM scores. This variant, denoted \emph{MVR-HandCamera}, uses the same RL and reward-model hyperparameters as MVR; the only change is that videos can also be rendered from a camera rigidly attached to the end-effector to better capture near-field contacts.

\Cref{tab:metaworld_wrist_camera} compares success rates for TQC, exocentric MVR, and MVR-HandCamera on the ten CW10 tasks. MVR-HandCamera achieves comparable average performance to MVR ($0.51 \pm 0.15$ vs.\ $0.54 \pm 0.22$), with clear gains on contact-rich tasks such as \textit{push-wall} and \textit{peg-unplug-side}, while slightly underperforming on some tasks that mainly require coarse arm motion (e.g., \textit{push-back}). These results suggest that near-field wrist views are particularly beneficial when local object–hand interactions determine success, whereas they may provide limited benefit or mild distraction on tasks dominated by global motion.

\begin{table}[htbp]
\centering
\caption{\textbf{MetaWorld success rates with a wrist-mounted camera.} TQC uses only task rewards. MVR uses exocentric videos, while MVR-HandCamera additionally incorporates a wrist-mounted hand camera in the reward pipeline.}
\label{tab:metaworld_wrist_camera}
\begin{tabular}{lccc}
\toprule
\textbf{Task} & \textbf{TQC} & \textbf{MVR} & \textbf{MVR-HandCamera} \\
\midrule
Hammer            & $0.20 \pm 0.40$ & $0.20 \pm 0.24$ & $0.20 \pm 0.24$ \\
Push-Wall         & $0.10 \pm 0.20$ & $0.60 \pm 0.49$ & $0.80 \pm 0.40$ \\
Faucet-Close      & $1.00 \pm 0.00$ & $1.00 \pm 0.00$ & $1.00 \pm 0.00$ \\
Push-Back         & $0.60 \pm 0.37$ & $0.50 \pm 0.45$ & $0.20 \pm 0.24$ \\
Stick-Pull        & $0.10 \pm 0.20$ & $0.20 \pm 0.24$ & $0.00 \pm 0.00$ \\
Handle-Press-Side & $1.00 \pm 0.00$ & $1.00 \pm 0.00$ & $1.00 \pm 0.00$ \\
Push              & $0.10 \pm 0.20$ & $0.10 \pm 0.20$ & $0.10 \pm 0.20$ \\
Shelf-Place       & $0.00 \pm 0.00$ & $0.20 \pm 0.24$ & $0.00 \pm 0.00$ \\
Window-Close      & $1.00 \pm 0.00$ & $1.00 \pm 0.00$ & $1.00 \pm 0.00$ \\
Peg-Unplug-Side   & $0.30 \pm 0.40$ & $0.60 \pm 0.37$ & $0.80 \pm 0.40$ \\
\midrule
Average           & $0.44 \pm 0.18$ & $0.54 \pm 0.22$ & $0.51 \pm 0.15$ \\
\bottomrule
\end{tabular}
\end{table}

\subsection{MetaWorld Training Curves}
\label{subsec:metaworld_training_curves}

To further examine how MVR shapes the learning dynamics, we report success rate as a function of training steps for two MetaWorld tasks where the performance gap between MVR and TQC is particularly pronounced: \textit{push-wall} and \textit{peg-unplug-side}.
In both cases, the MVR curves lie above the TQC baseline already in the early training phase, indicating that the shaped visual reward accelerates learning rather than only improving final asymptotic performance.

\begin{figure}[htbp]
\centering
\begin{subfigure}[b]{0.48\linewidth}
    \centering
    \includegraphics[width=\linewidth]{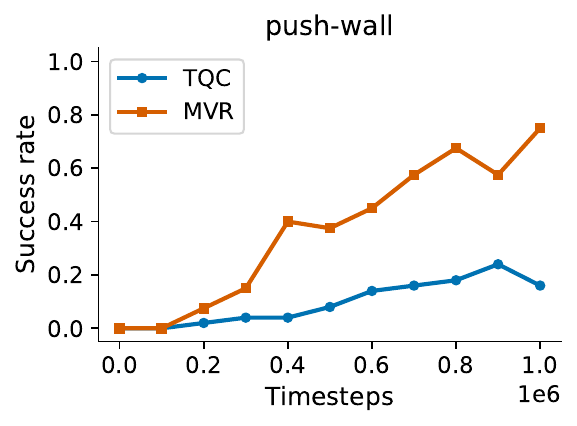}
    \caption{\textit{push-wall}.}
    \label{fig:metaworld_push_wall_curve}
\end{subfigure}
\hfill
\begin{subfigure}[b]{0.48\linewidth}
    \centering
    \includegraphics[width=\linewidth]{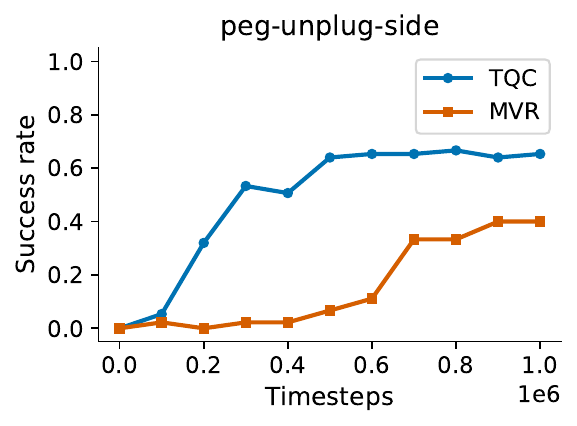}
    \caption{\textit{peg-unplug-side}.}
    \label{fig:metaworld_peg_unplug_curve}
\end{subfigure}
\caption{\textbf{Success rate vs.\ training step on two MetaWorld tasks.}
Across both (a) \textit{push-wall} and (b) \textit{peg-unplug-side}, MVR consistently outperforms TQC throughout training, with success rates already higher in the early stages, indicating that multi-view visual shaping accelerates learning rather than only improving final performance.}
\label{fig:metaworld_training_curves}
\end{figure}

\subsection{Additional Ablations}
\label{subsec:additional_ablations}

\begin{table}[htbp]
    \centering
    \caption{\textbf{Ablating the reward components.} Using only the VLM reward fails to solve the humanoid tasks, whereas combining it with $r^{task}$ recovers the full performance reported in \cref{Performance}.}
    \label{tab:rvlm_only}
    \begin{tabular}{lcccc}
        \toprule
        Method & Stand & Run & Slide & Sit\_Hard \\
        \midrule
        MVR (only $r^\text{VLM}$) & $7.73 \pm 0.68$ & $1.51 \pm 0.05$ & $0.83 \pm 0.02$ & $2.60 \pm 0.19$ \\
        MVR (full) & $918.55 \pm 29.30$ & $749.23 \pm 56.82$ & $735.03 \pm 142.85$ & $756.67 \pm 108.79$ \\
        TQC (only $r^{task}$) & $576.59 \pm 371.00$ & $647.87 \pm 186.98$ & $514.91 \pm 106.36$ & $511.85 \pm 155.45$ \\
        \bottomrule
    \end{tabular}
\end{table}

The comparison shows that relying exclusively on $r^\text{VLM}$ collapses learning: returns stay below ten across all four tasks. Adding the standard task reward lifts performance back to the levels reported in the main paper, highlighting that the VLM signal acts as a complement rather than a replacement for $r^{task}$.

\begin{table}[htbp]
    \centering
    \caption{\textbf{Temporal averaging versus attention-based pooling.} Attention can help static tasks but remains unstable on dynamic locomotion tasks, so we adopt simple averaging in the main paper.}
    \label{tab:attention_pooling}
    \begin{tabular}{lcccc}
        \toprule
        Task & MVR (Temporal Avg.) & MVR (Attention) \\
        \midrule
        Stand & $918.55 \pm 29.30$ & $940.80 \pm 11.11$ \\
        Run & $749.23 \pm 56.82$ & $519.83 \pm 366.45$ \\
        Slide & $735.03 \pm 142.85$ & $683.57 \pm 243.58$ \\
        Sit\_Hard & $756.67 \pm 108.79$ & $822.27 \pm 18.04$ \\
        \bottomrule
    \end{tabular}
\end{table}

For the pooling study, attention improves static, single-stage behaviours such as Stand and Sit\_Hard but degrades performance on locomotion tasks (Run, Slide) where the policy must react continuously. Because the weighting network observes only individual states, its outputs become noisy when the task depends on long-horizon momentum, reinforcing our choice of temporal averaging in the main ablations.

\paragraph{All-view VLM ablation.} To ablate the rendering strategy, we evaluate an all-view variant, \emph{MVR-ALL}, which renders four synchronized videos per rollout and concatenates them before querying the VLM so that every viewpoint is processed jointly. This eliminates viewpoint selection but multiplies rendering and encoding cost by four. \Cref{tab:mvr_all_views_appendix} compares TQC, MVR-ALL, and our default MVR. The results are informative: MVR-ALL delivers a noticeable uplift on the static `Stand` task yet offers only marginal or no gains on the dynamic locomotion tasks. For high-speed motions, maintaining a single coherent viewpoint preserves temporal consistency that the naive multi-view concatenation tends to disrupt, which explains why our lightweight single-view sampling coupled with relevance aggregation remains more effective despite the reduced rendering cost.

\begin{table}[h]
\scriptsize
\centering
\caption{\textbf{HumanoidBench returns for the all-view baseline.} MVR-ALL renders four videos simultaneously and feeds them jointly to the VLM. Means and standard deviations are computed over three seeds.}
\label{tab:mvr_all_views_appendix}
\begin{tabular}{lccc}
\toprule
Task & TQC & MVR-ALL & MVR \\
\midrule
Stand & $576.59 \pm 371.00$ & $879.06 \pm 54.27$ & $\mathbf{918.55 \pm 29.30}$ \\
Run & $647.87 \pm 186.98$ & $672.53 \pm 115.21$ & $\mathbf{749.23 \pm 56.82}$ \\
Slide & $514.91 \pm 106.36$ & $596.27 \pm 59.93$ & $\mathbf{735.03 \pm 142.85}$ \\
Sit\_Hard & $511.85 \pm 155.45$ & $510.78 \pm 115.49$ & $\mathbf{756.67 \pm 108.79}$ \\
\bottomrule
\end{tabular}
\end{table}

\subsection{Reward Correlation Analysis}
\label{subsec:reward_alignment_app}

\textit{Quantitative alignment.} We evaluated 100 Sit\_Hard rollouts and computed the trajectory-averaged $f^\text{VLM}$, $f^\text{MVR}$, and $r^\text{MVR}$. As summarised in \Cref{tab:reward_correlation_app}, their Pearson correlations with the binary success indicator reach $0.91$, $0.96$, and $0.98$, respectively. Importantly, the relevance score $f^\text{MVR}$ keeps a strong link to success while remaining only weakly correlated with the sparse environment reward ($0.22$), indicating that the learned shaping emphasises task completion cues rather than copying $r^{task}$. Consequently the shaped reward $r^\text{MVR}$ maintains the success alignment and, because it respects the ordering of successful trajectories, leaves the optimal policy approximately unchanged while reinforcing the visual priors encoded by the relevance model. These measurements explain why visual feedback is indispensable: it resolves ambiguities in $r^{task}$ while remaining tightly aligned with the actual success criterion.

\textit{Automatic shaping.} Tracking the same 100 rollouts over training shows that the average magnitude of $r^\text{VLM}$ during the final 100 steps of successful episodes drops to about $23\%$ of its early-training value. This confirms empirically that $r^\text{MVR}$ hands control back to $r^{task}$ as soon as the desired behaviour emerges, preventing long-term gradient conflicts.

\Cref{tab:reward_correlation_app} summarises the correlations measured on Sit\_Hard (identical to those referenced in the main text). The statistics highlight that $f^{\text{MVR}}$ remains strongly aligned with success yet only weakly correlated with $r^{task}$, while the shaped reward $r^{\text{MVR}}$ maintains the ordering of successful rollouts, keeping the optimal policy approximately unchanged.

\begin{table}[h]
\scriptsize
\centering
\caption{\textbf{Reward-signal correlations on Sit\_Hard.}}
\label{tab:reward_correlation_app}
\begin{tabular}{lcc}
\toprule
\textbf{Signal} & \textbf{Pearson vs success} & \textbf{Pearson vs $r^{task}$} \\
\midrule
Raw VLM similarity $f^{\text{VLM}}$ & 0.91 & 0.21 \\
Learned relevance $f^{\text{MVR}}$ & 0.96 & 0.22 \\
Shaped reward $r^{\text{MVR}}$ & 0.98 & 0.85 \\
\bottomrule
\end{tabular}
\end{table}

We further compare reward quality across VLM backbones. \Cref{tab:vlm_reward_quality_app} reports the correlation between the raw similarity signal and $r^{task}$ together with the fraction of trajectories exhibiting negative correlation; the raw signal varies widely across ViCLIP-L and ViCLIP-B, whereas the learned relevance remains small but positive in both cases.

We also assess robustness with respect to success correlation. \Cref{tab:vlm_corr_appendix} shows that the learned relevance and shaped reward remain positive and nearly unchanged across the two backbones, so the MVR pipeline still produces a stable, task-aligned signal when the base VLM is reduced in capacity.

\begin{table}[h]
\scriptsize
\centering
\caption{\textbf{Correlation with $r^{task}$ and fraction of negatively correlated trajectories (Sit\_Hard).}}
\label{tab:vlm_reward_quality_app}
\begin{tabular}{lccc}
\toprule
\textbf{Backbone} & $\text{corr}(f^{\text{VLM}}, r^{task})$ & $\text{Negative traj.}$ & $\text{corr}(f^{\text{MVR}}, r^{task})$ \\
\midrule
ViCLIP-L (428M) & 0.21 & 50\% & 0.22 \\
ViCLIP-B (150M) & 0.51 & 10\% & 0.23 \\
\bottomrule
\end{tabular}
\end{table}

\begin{table}[h]
\scriptsize
\centering
\caption{\textbf{Correlation with success across VLM backbones (Sit\_Hard).}}
\label{tab:vlm_corr_appendix}
\begin{tabular}{lccc}
\toprule
\textbf{Backbone} & $\text{corr}(f^{\text{VLM}},\,\text{success})$ & $\text{corr}(f^{\text{MVR}},\,\text{success})$ & $\text{corr}(r^{\text{MVR}},\,\text{success})$ \\
\midrule
ViCLIP-L (428M) & 0.91 & 0.96 & 0.98 \\
ViCLIP-B (150M) & 0.90 & 0.95 & 0.97 \\
\bottomrule
\end{tabular}
\end{table}

\subsection{The Effect of Weighting Parameter \texorpdfstring{$w$}{w}}
\label{subsec:weight}

\begin{figure}[htbp]
    \centering
    \includegraphics[width=0.8\linewidth]{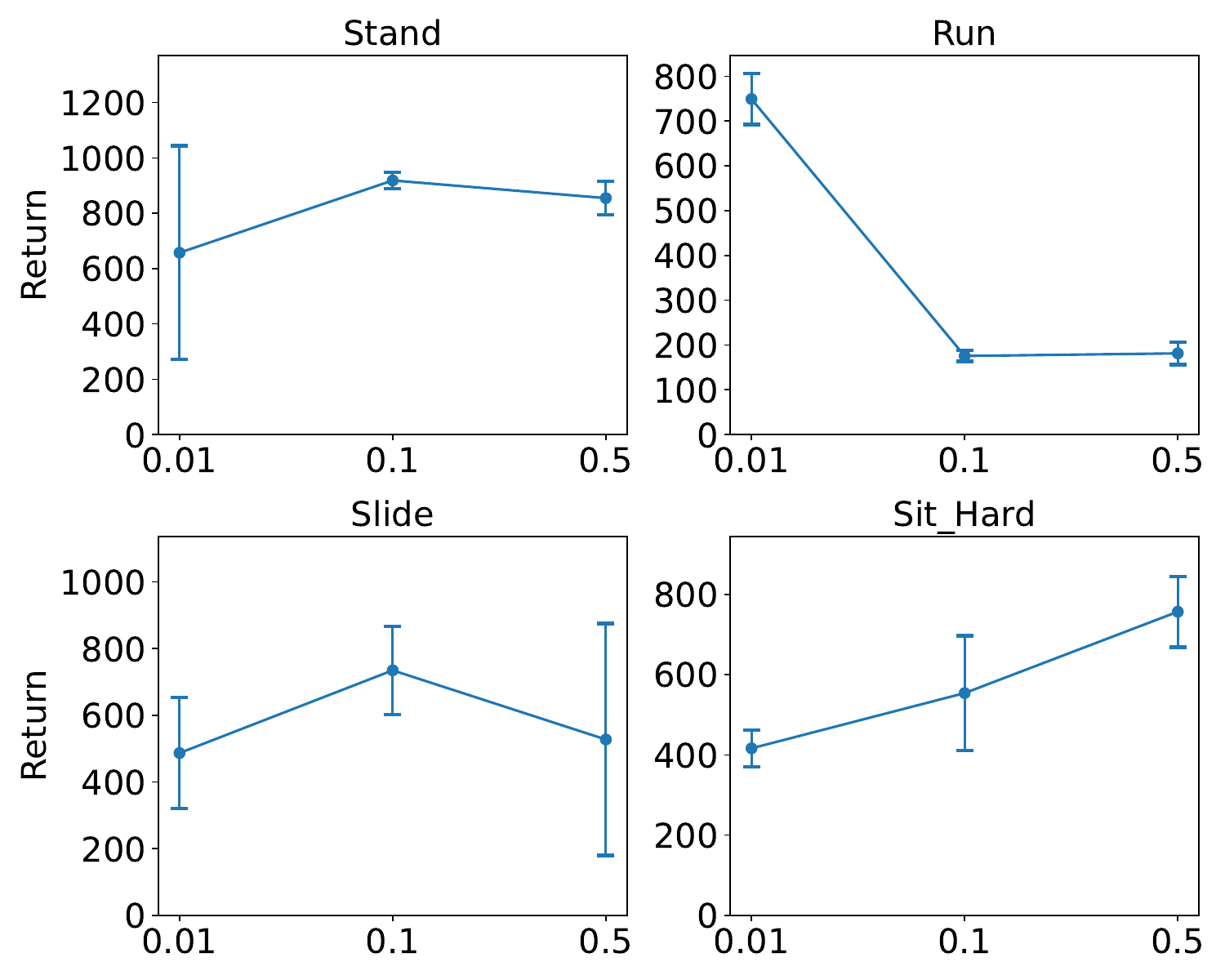}
    \caption{\textbf{The optimal value of $w$ in \cref{eq:policy_obj} is task-dependent}. Smaller values ($w=0.01$) appear beneficial for tasks requiring rapid motion like 'Run', allowing the agent to prioritize the task reward. Conversely, tasks emphasizing stability like 'Stand' benefit from larger values ($w=0.1$ or $w=0.5$), giving more weight to the VLM guidance. 'Sit\_Hard' shows peak performance at $w=0.5$, while 'Slide' performs best at $w=0.1$. This indicates that tuning $w$ based on task requirements may be necessary for optimal results.}
    \label{fig:hyperablation2}
\end{figure}

The MVR framework introduces a hyperparameter $w$ that scales the VLM-derived component in the shaped reward $r^{\text{MVR}}$ within the policy objective \cref{eq:policy_obj}. We investigated the impact of different values of $w$ on MVR's performance, selecting $w$ from $\{0.01, 0.1, 0.5\}$ via grid search. \Cref{fig:hyperablation2} shows the final performance for different $w$ values across selected tasks.

The results suggest that the optimal balance between task rewards and VLM guidance is task-specific. Tasks involving rapid, dynamic motion (e.g., 'Run') may benefit from smaller $w$, prioritizing the optimization of the primary task objective ($r^{task}$). Tasks requiring precise posture or stability (e.g., 'Stand') might benefit more from larger $w$, leveraging the VLM's visual understanding encoded in $r^{\text{MVR}}$ to achieve the desired configuration. This highlights the need for careful tuning of $w$ according to the specific demands of the task. \citep{hu2020learning} is a potential solution for automatic parameter selection.

\subsection{Comparison with FuRL}
\label{subsec:comparison_furl}

We include additional comparative results against FuRL~\citep{fu2024furl}, a method that also utilizes visual information for reward generation. FuRL is based on image-text similarity, and its core mechanism involves leveraging successful trajectories to perform online fine-tuning of the Vision-Language Model's (VLM) representations. This online fine-tuning aims to address potential VLM misalignment with specific task objectives. To facilitate a comparison on HumanoidBench, we operated FuRL under a VLM interaction and representation update schedule comparable to MVR's (i.e., fitting a reward model based on a similar rendering frequency), ensuring computational tractability for these demanding tasks. The results for FuRL on MetaWorld tasks, as presented in \cref{tab:metaworld_results_furl}, are taken directly from the original FuRL publication~\citep{fu2024furl}. For HumanoidBench, we conducted new experiments with FuRL operating under this adjusted interaction scheme, and these results are summarized in \cref{tab:performance_furl_humanoid}.

FuRL's performance on HumanoidBench tasks (Table \ref{tab:performance_furl_humanoid}), when operated under a VLM interaction schedule adjusted for scalability, is notably lower. A core potential reason is that FuRL's strategy of fine-tuning VLM representations using successful trajectories is severely hampered in HumanoidBench. The inherent difficulty of exploration in these complex environments (cf. \Cref{subsec:algorithms}) likely means that high-quality successful trajectories, which are crucial for FuRL's VLM fine-tuning process, are rarely discovered. This scarcity of effective training data would directly impede the optimization of VLM embeddings, thus limiting the quality of the visual guidance FuRL can provide for these challenging humanoid control tasks.

% MetaWorld results table
\begin{table}[h]
\scriptsize
\centering
\caption{\textbf{MVR demonstrates competitive or superior performance compared to FuRL on the selected MetaWorld tasks for which FuRL results are available.} Both methods combine task-specific rewards ($r^\text{task}$) with VLM-based rewards. FuRL results are sourced from~\citep{fu2024furl}.}
\label{tab:metaworld_results_furl}
\begin{tabular}{lcc}
\toprule
Task & MVR & FuRL \\
\midrule
push & $\mathbf{0.10 \pm 0.20}$ & $0.06 \pm 0.08$ \\
window-close & $\mathbf{1.00 \pm 0.00}$ & $\mathbf{1.00 \pm 0.00}$ \\
\midrule
\textbf{Average} & $\mathbf{0.55 \pm 0.10}$ & $0.53 \pm 0.04$ \\
\bottomrule
\end{tabular}
\end{table}

% HumanoidBench results table
\begin{table*}[h]
\centering
\scriptsize
\caption{\textbf{MVR significantly outperforms FuRL on all tested dynamic and static HumanoidBench tasks.} For this comparison, FuRL, which combines task rewards ($r^\text{task}$) with VLM-based rewards, was assessed using a VLM guidance mechanism updated at a frequency similar to MVR's for scalability. These results further underscore the effectiveness of MVR's approach in leveraging multi-view video guidance from VLMs for complex humanoid control.}
\label{tab:performance_furl_humanoid}
\vskip 0.1in
% \resizebox{\textwidth}{!}{%
\begin{tabular}{lcc}
\toprule
Task & \ac{MVR} & FuRL\\
\midrule
Walk            & $\mathbf{927.47 \pm 1.83}$  & $1.64 \pm 0.31$   \\
Run             & $\mathbf{749.23 \pm 56.82}$ & $13.32 \pm 12.79$ \\
Stair           & $\mathbf{208.60 \pm 166.22}$ & $4.41 \pm 2.32$  \\
Slide           & $\mathbf{735.03 \pm 142.85}$ & $19.74 \pm 11.83$ \\
\midrule
Stand           & $\mathbf{918.55 \pm 29.30}$ & $30.15 \pm 8.31$   \\
Sit\_Simple     & $\mathbf{861.07 \pm 19.35}$ & $196.15 \pm 154.17$ \\
Sit\_Hard       & $\mathbf{756.67 \pm 108.79}$ & $24.56 \pm 2.69$  \\
Balance\_Simple & $\mathbf{299.87 \pm 58.45}$ & $21.32 \pm 11.43$ \\
Balance\_Hard   & $\mathbf{95.78 \pm 11.03}$  & $15.67 \pm 3.11$  \\
\bottomrule
\end{tabular}
% }
\end{table*}

\subsection{Performance against Original RoboCLIP on HumanoidBench}
We further benchmark MVR against the original RoboCLIP codebase (PPO, S3D) on HumanoidBench. This specific evaluation is critical as it assesses MVR against RoboCLIP in its unaltered, publicly available form. The substantially poorer performance of the original RoboCLIP, detailed in Table \ref{tab:performance_mvr_vs_roboclip_furl_mvr_data}, likely stems from several factors inherent to its design. Firstly, its reliance on the S3D vision model, while capable of video processing, may provide less nuanced visual-semantic understanding for reward generation compared to the more advanced VLMs leveraged by MVR. Secondly, RoboCLIP's original reward mechanism might not as effectively capture complex, temporally extended behaviors or handle viewpoint limitations as MVR's multi-view video-based approach. This contrasts with other analyses in this paper where baseline configurations (e.g., PPO to TQC, S3D to ViCLIP) were adapted for fairer component-wise comparisons, isolating specific architectural contributions.

% HumanoidBench results table - MVR vs RoboCLIP with MVR data from FuRL comparison
\begin{table*}[h]
\centering
\scriptsize
\caption{\textbf{MVR substantially outperforms the original RoboCLIP (PPO, S3D) across all HumanoidBench tasks.} The original RoboCLIP combines task rewards ($r^\text{task}$) with VLM-based rewards from its S3D model.}
\label{tab:performance_mvr_vs_roboclip_furl_mvr_data}
\vskip 0.1in
% \resizebox{\textwidth}{!}{%
\begin{tabular}{llcc}
\toprule
Category & Task & \ac{MVR} & RoboCLIP \\
\midrule
\multirow{4}{*}{Dynamic} & Walk           & $\mathbf{927.47 \pm 1.83}$   & $25.31 \pm 5.98$  \\
                         & Run            & $\mathbf{581.33 \pm 210.35}$ & $8.19 \pm 3.09$   \\
                         & Stair          & $\mathbf{164.74 \pm 132.96}$ & $19.25 \pm 8.63$  \\
                         & Slide          & $\mathbf{622.05 \pm 195.11}$ & $27.37 \pm 8.28$  \\
\midrule
\multirow{5}{*}{Static}  & Stand          & $\mathbf{925.55 \pm 25.10}$  & $36.67 \pm 7.66$  \\
                         & Sit\_Simple    & $\mathbf{757.34 \pm 143.96}$ & $16.95 \pm 1.10$ \\
                         & Sit\_Hard      & $\mathbf{705.29 \pm 105.18}$ & $11.53 \pm 4.20$  \\
                         & Balance\_Simple & $\mathbf{284.54 \pm 46.82}$  & $45.51 \pm 1.39$  \\
                         & Balance\_Hard   & $\mathbf{95.78 \pm 11.03}$   & $36.42 \pm 1.83$  \\
\bottomrule
\end{tabular}
% }
\end{table*}

\subsection{Pixel-Based MetaWorld Experiments}
We additionally evaluate MVR on the MetaWorld suite when policies receive pixel observations. Each agent augments proprioceptive input with embeddings from the R3M-ViT encoder~\citep{nair2022r3m}, while MVR reuses the multi-view reward pipeline without extra tuning. The shaped reward $r_t^{\text{MVR}}$ still combines environment feedback with the learned visual guidance. Results in \cref{tab:metaworld_pixel_followup} show that MVR preserves its success-rate advantage even with raw RGB inputs.

\begin{table}[h]
\scriptsize
\centering
\caption{\textbf{Pixel-conditioned MetaWorld success rates.} Both agents operate on proprioception concatenated with R3M-ViT features. Returns are averaged over five seeds.}
\label{tab:metaworld_pixel_followup}
\begin{tabular}{lcc}
\toprule
Task & TQC (vision) & MVR (vision) \\
\midrule
hammer & $0.17 \pm 0.24$ & $\mathbf{0.80 \pm 0.28}$ \\
push-wall & $\mathbf{0.33 \pm 0.47}$ & $0.17 \pm 0.24$ \\
faucet-close & $\mathbf{1.00 \pm 0.00}$ & $\mathbf{1.00 \pm 0.00}$ \\
push-back & $0.00 \pm 0.00$ & $0.00 \pm 0.00$ \\
stick-pull & $0.00 \pm 0.00$ & $0.00 \pm 0.00$ \\
handle-press-side & $\mathbf{1.00 \pm 0.00}$ & $\mathbf{1.00 \pm 0.00}$ \\
push & $0.00 \pm 0.00$ & $0.00 \pm 0.00$ \\
shelf-place & $0.00 \pm 0.00$ & $0.00 \pm 0.00$ \\
window-close & $\mathbf{1.00 \pm 0.00}$ & $\mathbf{1.00 \pm 0.00}$ \\
peg-unplug-side & $0.00 \pm 0.00$ & $\mathbf{0.20 \pm 0.22}$ \\
\midrule
Average & $0.35 \pm 0.07$ & $\mathbf{0.42 \pm 0.07}$ \\
\bottomrule
\end{tabular}
\end{table}

\subsection{Sparse-Reward MetaWorld Experiments}
\label{subsec:sparse_metaworld}

To further assess MVR in settings where dense task rewards are unavailable, we construct a sparse-reward variant of the MetaWorld benchmark.
In this setting, we remove all dense environment rewards and keep only the binary success signal, while leaving the visual shaping signal $r^{\text{VLM}}$ unchanged.
We then compare three configurations: TQC with dense rewards only (TQC-Dense), MVR with dense rewards (MVR-Dense, identical to the main-paper setting), and MVR with sparse environment rewards (MVR-Sparse).

\begin{table}[h]
\scriptsize
\centering
\caption{\textbf{Sparse-reward MetaWorld experiments.} MVR-Sparse uses only a sparse success signal from the environment plus visual shaping, yet achieves competitive performance with the dense-reward baseline TQC-Dense. Results are averaged over five seeds.}
\label{tab:metaworld_sparse}
\resizebox{0.8\textwidth}{!}{%
\begin{tabular}{lccc}
\toprule
Task & TQC-Dense & MVR-Dense & MVR-Sparse \\
\midrule
hammer            & $0.20 \pm 0.40$ & $0.20 \pm 0.24$ & $\mathbf{0.90 \pm 0.20}$ \\
push-wall         & $0.10 \pm 0.20$ & $\mathbf{0.60 \pm 0.49}$ & $0.30 \pm 0.24$ \\
faucet-close      & $\mathbf{1.00 \pm 0.00}$ & $\mathbf{1.00 \pm 0.00}$ & $\mathbf{1.00 \pm 0.00}$ \\
push-back         & $\mathbf{0.60 \pm 0.37}$ & $0.50 \pm 0.45$ & $0.00 \pm 0.00$ \\
stick-pull        & $0.10 \pm 0.20$ & $\mathbf{0.20 \pm 0.24}$ & $0.00 \pm 0.00$ \\
handle-press-side & $\mathbf{1.00 \pm 0.00}$ & $\mathbf{1.00 \pm 0.00}$ & $\mathbf{1.00 \pm 0.00}$ \\
push              & $\mathbf{0.10 \pm 0.20}$ & $\mathbf{0.10 \pm 0.20}$ & $0.00 \pm 0.00$ \\
shelf-place       & $0.00 \pm 0.00$ & $\mathbf{0.20 \pm 0.24}$ & $0.00 \pm 0.00$ \\
window-close      & $\mathbf{1.00 \pm 0.00}$ & $\mathbf{1.00 \pm 0.00}$ & $\mathbf{1.00 \pm 0.00}$ \\
peg-unplug-side   & $0.30 \pm 0.40$ & $0.60 \pm 0.37$ & $\mathbf{0.90 \pm 0.20}$ \\
\midrule
Average           & $0.44 \pm 0.18$ & $\mathbf{0.54 \pm 0.22}$ & $0.51 \pm 0.064$ \\
\bottomrule
\end{tabular}%
}
\end{table}

Overall, MVR-Sparse attains an average success rate comparable to TQC-Dense while removing dense environment rewards, indicating that MVR can leverage visual shaping to compensate for the lack of hand-engineered dense task rewards.

\section{Additional Illustration}

\subsection{Objective Functions for State Relevance Learning}
\label{subsec:learning_relevance}

\begin{figure}[htbp] % Changed from [t] to [htbp] for more flexibility
  \centering
  \includegraphics[width=0.8\linewidth]{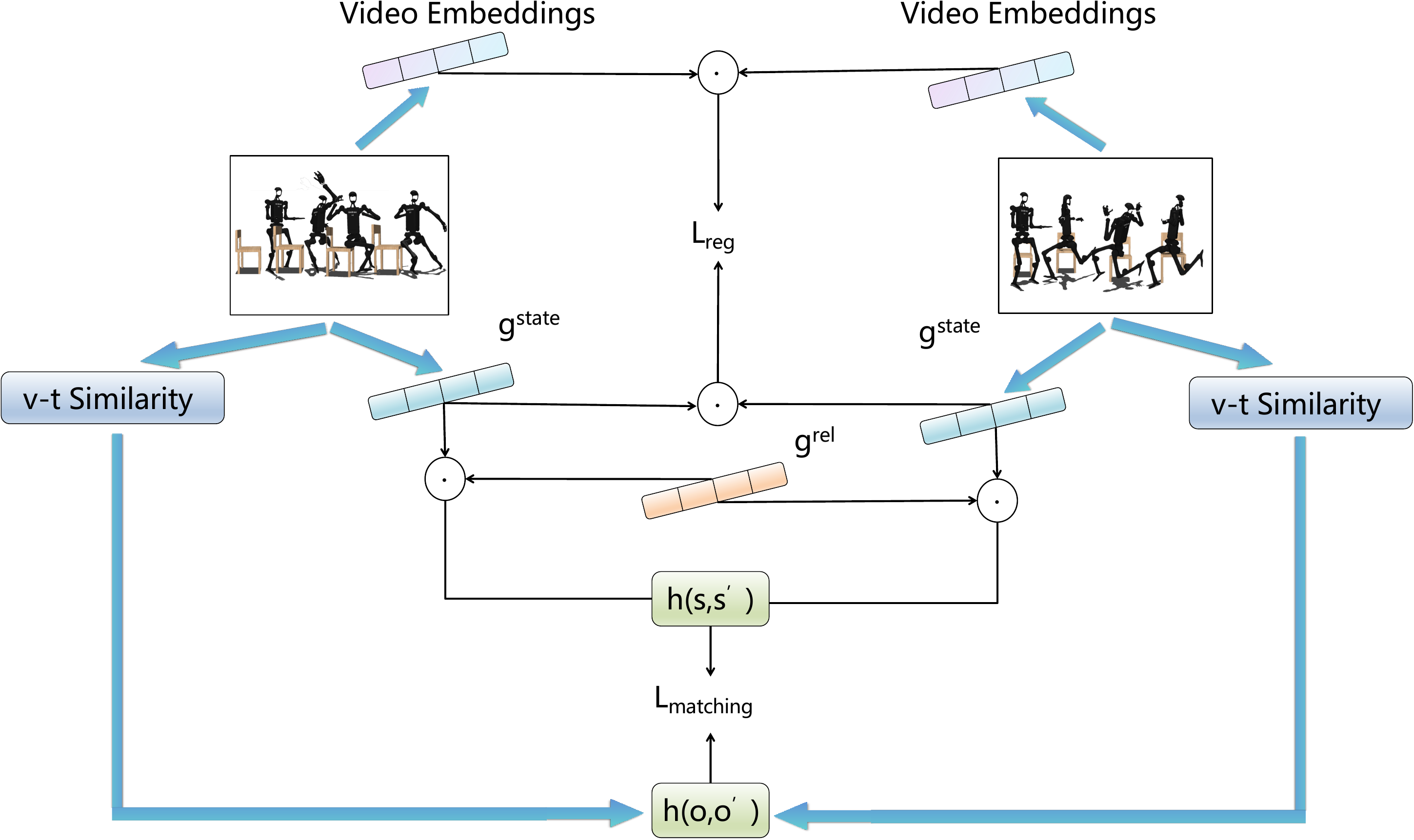}
    \caption{\textbf{The two objectives for learning state relevance}. The state relevance function $f^\text{MVR}$ comprises a state encoder $g^\text{state}$ and a relevance vector $g^\text{rel}$. $L_\text{matching}$ aligns the relative ranking of state sequences (derived from $f^\text{MVR}$) with the ranking induced by VLM video-text similarity scores. $L_\text{reg}$ regularizes the state sequence representations by enforcing consistency with the similarity structure observed in the VLM's video embedding space. This dual approach allows MVR to learn state relevance effectively while mitigating noise from visual details and viewpoint variations.}
    \label{fig:loss}
\vskip -0.1in % Kept from original if vertical spacing adjustment is desired
\end{figure}

As detailed in \cref{subsec:relevance_learning}, learning the state relevance function $f^{\text{MVR}}$ involves addressing challenges arising from the discrepancy between state-space representations and visual video data, as well as viewpoint-induced variability in similarity scores when the same trajectory is rendered from different cameras. MVR tackles these using two learning objectives, $L_{\text{matching}}$ and $L_{\text{reg}}$, illustrated in \Cref{fig:loss}.

$L_{\text{matching}}$ encourages the state relevance function to preserve the relative ordering of behaviors indicated by the VLM's assessment of corresponding videos. $L_{\text{reg}}$ aligns the distance metric in the learned state embedding space with the distances in the VLM's video embedding space, promoting representations robust to view-specific interference. Together, these objectives enable MVR to learn a meaningful state relevance function from multi-view video feedback.

\section{Limitations and Future Work}
\label{sec:limitations}

MVR currently relies on a single textual description per task, which can underspecify complex behaviours. Extending the framework with adaptive or stage-specific prompts is a promising direction for capturing richer task semantics.

The prototype assumes access to multiple calibrated cameras to gather multi-view footage. Deployments with limited viewpoints may require coupling MVR with novel-view synthesis modules such as NeRFs~\citep{mildenhall2020nerf} or active camera placement to maintain coverage.

While VLM guidance helps expose dynamic motion cues, combining video-text similarity with other perceptual signals may broaden applicability to tasks where text alignment alone is insufficient. Finally, automating the selection of the weighting parameter $w$ in the policy objective would reduce hyperparameter tuning effort and improve robustness across domains.

\section{Broader Impacts}
\label{sec:broader_impacts}
Robot skill learning is an important topic for robotics and embodied artificial intelligence. This paper presents an approach that solicits feedback from VLMs for skill learning, thereby leveraging the rich knowledge stored in such models. With such a method, this paper aims to pave the road towards automatic robot skill learning, a desideratum of learning systems that enables intelligent robots to acquire diverse real-world tasks.

As a foundational methodological contribution, this work does not present direct applications with immediately foreseeable negative societal impacts. Nevertheless, continued research in this foundational area, conducted with an awareness of broader ethical considerations, is crucial for building a deeper understanding and fostering the responsible development of more advanced autonomous systems in the future.

\paragraph{LLM Usage.} A general-purpose large language model (ChatGPT) was used solely for polishing grammar and improving the clarity of draft paragraphs. All technical content, equations, experimental design, and analysis were created and verified by the authors.\par\medskip

%% file: main.bbl
\begin{thebibliography}{53}
\providecommand{\natexlab}[1]{#1}
\providecommand{\url}[1]{\texttt{#1}}
\expandafter\ifx\csname urlstyle\endcsname\relax
  \providecommand{\doi}[1]{doi: #1}\else
  \providecommand{\doi}{doi: \begingroup \urlstyle{rm}\Url}\fi

\bibitem[Alakuijala et~al.(2025)Alakuijala, McLean, Woungang, Kaski, Marttinen,
  Yuan, and Farsad]{alakuijala2025vlc}
Minttu Alakuijala, Reginald McLean, Isaac Woungang, Samuel Kaski, Pekka
  Marttinen, Kai Yuan, and Nariman Farsad.
\newblock Video-language critic: Transferable reward functions for
  language-conditioned robotics.
\newblock \emph{Transactions on Machine Learning Research}, 2025.

\bibitem[Ayalew et~al.(2024)Ayalew, Zhang, Wu, Jiang, Maire, and
  Walter]{arxiv_2411.17764}
Tewodros~W. Ayalew, Xiao Zhang, Kevin~Yuanbo Wu, Tianchong Jiang, Michael
  Maire, and Matthew~R. Walter.
\newblock Progressor: A perceptually guided reward estimator with
  self-supervised online refinement.
\newblock \emph{arXiv preprint arXiv:2411.17764}, 2024.

\bibitem[Bradley \& Terry(1952)Bradley and Terry]{10.2307/2334029}
Ralph~Allan Bradley and Milton~E. Terry.
\newblock Rank analysis of incomplete block designs: I. the method of paired
  comparisons.
\newblock \emph{Biometrika}, 39\penalty0 (3/4):\penalty0 324--345, 1952.
\newblock ISSN 00063444.

\bibitem[brian ichter et~al.(2022)brian ichter, Brohan, Chebotar, Finn,
  Hausman, Herzog, Ho, Ibarz, Irpan, Jang, Julian, Kalashnikov, Levine, Lu,
  Parada, Rao, Sermanet, Toshev, Vanhoucke, Xia, Xiao, Xu, Yan, Brown, Ahn,
  Cortes, Sievers, Tan, Xu, Reyes, Rettinghouse, Quiambao, Pastor, Luu, Lee,
  Kuang, Jesmonth, Jeffrey, Ruano, Hsu, Gopalakrishnan, David, Zeng, and
  Fu]{ichter2022do}
brian ichter, Anthony Brohan, Yevgen Chebotar, Chelsea Finn, Karol Hausman,
  Alexander Herzog, Daniel Ho, Julian Ibarz, Alex Irpan, Eric Jang, Ryan
  Julian, Dmitry Kalashnikov, Sergey Levine, Yao Lu, Carolina Parada, Kanishka
  Rao, Pierre Sermanet, Alexander~T Toshev, Vincent Vanhoucke, Fei Xia, Ted
  Xiao, Peng Xu, Mengyuan Yan, Noah Brown, Michael Ahn, Omar Cortes, Nicolas
  Sievers, Clayton Tan, Sichun Xu, Diego Reyes, Jarek Rettinghouse, Jornell
  Quiambao, Peter Pastor, Linda Luu, Kuang-Huei Lee, Yuheng Kuang, Sally
  Jesmonth, Kyle Jeffrey, Rosario~Jauregui Ruano, Jasmine Hsu, Keerthana
  Gopalakrishnan, Byron David, Andy Zeng, and Chuyuan~Kelly Fu.
\newblock Do as i can, not as i say: Grounding language in robotic affordances.
\newblock In \emph{6th Annual Conference on Robot Learning}, 2022.

\bibitem[Chan et~al.(2023)Chan, Mnih, Behbahani, Laskin, Wang, Pardo, Gazeau,
  Sahni, Horgan, Baumli, Schroecker, Spencer, Steigerwald, Quan, Comanici,
  Flennerhag, Neitz, Zhang, Schaul, Singh, Lyle, Rockt{\"a}schel,
  Parker-Holder, and Holsheimer]{baumli2023vision}
Harris Chan, Volodymyr Mnih, Feryal Behbahani, Michael Laskin, Luyu Wang, Fabio
  Pardo, Maxime Gazeau, Himanshu Sahni, Dan Horgan, Kate Baumli, Yannick
  Schroecker, Stephen Spencer, Richie Steigerwald, John Quan, Gheorghe
  Comanici, Sebastian Flennerhag, Alexander Neitz, Lei~M Zhang, Tom Schaul,
  Satinder Singh, Clare Lyle, Tim Rockt{\"a}schel, Jack Parker-Holder, and
  Kristian Holsheimer.
\newblock Vision-language models as a source of rewards.
\newblock 2023.

\bibitem[Chen et~al.(2024)Chen, Mees, Kumar, and Levine]{chen2024vision}
William Chen, Oier Mees, Aviral Kumar, and Sergey Levine.
\newblock Vision-language models provide promptable representations for
  reinforcement learning.
\newblock \emph{arXiv preprint arXiv:2402.02651}, 2024.

\bibitem[Christiano et~al.(2017)Christiano, Leike, Brown, Martic, Legg, and
  Amodei]{NIPS2017_d5e2c0ad}
Paul~F Christiano, Jan Leike, Tom Brown, Miljan Martic, Shane Legg, and Dario
  Amodei.
\newblock Deep reinforcement learning from human preferences.
\newblock In \emph{Advances in Neural Information Processing Systems}, pp.\
  4302--4310. Curran Associates, Inc., 2017.

\bibitem[Cui et~al.(2024)Cui, Liu, Liu, Yang, Zhu, and Huang]{cui2024anyskill}
Jieming Cui, Tengyu Liu, Nian Liu, Yaodong Yang, Yixin Zhu, and Siyuan Huang.
\newblock Anyskill: Learning open-vocabulary physical skill for interactive
  agents.
\newblock In \emph{Proceedings of the IEEE/CVF Conference on Computer Vision
  and Pattern Recognition}, 2024.

\bibitem[Du et~al.(2023{\natexlab{a}})Du, Konyushkova, Denil, Raju, Landon,
  Hill, de~Freitas, and Cabi]{pmlr-v232-du23b}
Yuqing Du, Ksenia Konyushkova, Misha Denil, Akhil Raju, Jessica Landon, Felix
  Hill, Nando de~Freitas, and Serkan Cabi.
\newblock Vision-language models as success detectors.
\newblock In \emph{Proceedings of The 2nd Conference on Lifelong Learning
  Agents}, 2023{\natexlab{a}}.

\bibitem[Du et~al.(2023{\natexlab{b}})Du, Watkins, Wang, Colas, Darrell,
  Abbeel, Gupta, and Andreas]{du2023guiding}
Yuqing Du, Olivia Watkins, Zihan Wang, C{\'e}dric Colas, Trevor Darrell, Pieter
  Abbeel, Abhishek Gupta, and Jacob Andreas.
\newblock Guiding pretraining in reinforcement learning with large language
  models.
\newblock In \emph{International Conference on Machine Learning},
  2023{\natexlab{b}}.

\bibitem[Escontrela et~al.(2023)Escontrela, Adeniji, Yan, Jain, Peng, Goldberg,
  Lee, Hafner, and Abbeel]{escontrela2023video}
Alejandro Escontrela, Ademi Adeniji, Wilson Yan, Ajay Jain, Xue~Bin Peng, Ken
  Goldberg, Youngwoon Lee, Danijar Hafner, and Pieter Abbeel.
\newblock Video prediction models as rewards for reinforcement learning.
\newblock In \emph{Thirty-seventh Conference on Neural Information Processing
  Systems}, 2023.

\bibitem[Fu et~al.(2024)Fu, Zhang, Wu, Xu, and Boulet]{fu2024furl}
Yuwei Fu, Haichao Zhang, Di~Wu, Wei Xu, and Benoit Boulet.
\newblock Fu{RL}: Visual-language models as fuzzy rewards for reinforcement
  learning.
\newblock In \emph{roceedings of the Forty-first International Conference on
  Machine Learning}, 2024.

\bibitem[Gao et~al.(2023)Gao, Hu, Xu, and Xu]{gao2023can}
Jialu Gao, Kaizhe Hu, Guowei Xu, and Huazhe Xu.
\newblock Can pre-trained text-to-image models generate visual goals for
  reinforcement learning?
\newblock \emph{Advances in Neural Information Processing Systems}, 2023.

\bibitem[Hafner et~al.(2023)Hafner, Pasukonis, Ba, and
  Lillicrap]{hafner2023dreamerv3}
Danijar Hafner, Jurgis Pasukonis, Jimmy Ba, and Timothy Lillicrap.
\newblock Mastering diverse domains through world models.
\newblock \emph{arXiv preprint arXiv:2301.04104}, 2023.

\bibitem[Han et~al.(2024)Han, Chen, Williams, and Ravichandar]{han2024cimer}
Yunhai Han, Zhenyang Chen, Kyle~A. Williams, and Harish Ravichandar.
\newblock Learning prehensile dexterity by imitating and emulating state-only
  observations.
\newblock \emph{IEEE Robotics and Automation Letters}, 2024.

\bibitem[Hansen et~al.(2024)Hansen, Su, and Wang]{hansen2024tdmpc2}
Nicklas Hansen, Hao Su, and Xiaolong Wang.
\newblock Td-mpc2: Scalable, robust world models for continuous control.
\newblock In \emph{International Conference on Learning Representations}, 2024.

\bibitem[Hu et~al.(2020)Hu, Wang, Jia, Wang, Chen, Hao, Wu, and
  Fan]{hu2020learning}
Yujing Hu, Weixun Wang, Hangtian Jia, Yixiang Wang, Yingfeng Chen, Jianye Hao,
  Feng Wu, and Changjie Fan.
\newblock Learning to utilize shaping rewards: A new approach of reward
  shaping.
\newblock \emph{Advances in Neural Information Processing Systems}, 2020.

\bibitem[Huang et~al.(2024{\natexlab{a}})Huang, Lipovetzky, and
  Cohn]{huang2024dark}
Sukai Huang, Nir Lipovetzky, and Trevor Cohn.
\newblock The dark side of rich rewards: Understanding and mitigating noise in
  vlm rewards.
\newblock \emph{arXiv preprint arXiv:2409.15922}, 2024{\natexlab{a}}.

\bibitem[Huang et~al.(2024{\natexlab{b}})Huang, Jiang, Ze, and
  Xu]{huang2024diffusionreward}
Tao Huang, Guangqi Jiang, Yanjie Ze, and Huazhe Xu.
\newblock Diffusion reward: Learning rewards via conditional video diffusion.
\newblock In \emph{European Conference on Computer Vision (ECCV)},
  2024{\natexlab{b}}.

\bibitem[Kawaharazuka et~al.(2024)Kawaharazuka, Matsushima, Gambardella, Guo,
  Paxton, and Zeng]{kawaharazuka2024real}
Kento Kawaharazuka, Tatsuya Matsushima, Andrew Gambardella, Jiaxian Guo, Chris
  Paxton, and Andy Zeng.
\newblock Real-world robot applications of foundation models: A review.
\newblock \emph{Advanced Robotics}, 2024.

\bibitem[Kim et~al.(2023)Kim, Seo, Liu, Lee, Shin, Lee, and
  Lee]{NEURIPS2023_aa933b5a}
Changyeon Kim, Younggyo Seo, Hao Liu, Lisa Lee, Jinwoo Shin, Honglak Lee, and
  Kimin Lee.
\newblock Guide your agent with adaptive multimodal rewards.
\newblock In \emph{Advances in Neural Information Processing Systems}, 2023.

\bibitem[Kim et~al.(2025)Kim, Heo, Lee, Lee, Shin, Lim, and Lee]{kim2025reds}
Changyeon Kim, Minho Heo, Doohyun Lee, Honglak Lee, Jinwoo Shin, Joseph~J. Lim,
  and Kimin Lee.
\newblock {REDS}: Subtask-aware visual reward learning from segmented
  demonstrations.
\newblock In \emph{International Conference on Learning Representations}, 2025.

\bibitem[Kuznetsov et~al.(2020)Kuznetsov, Shvechikov, Grishin, and
  Vetrov]{kuznetsov2020controlling}
Arsenii Kuznetsov, Pavel Shvechikov, Alexander Grishin, and Dmitry Vetrov.
\newblock Controlling overestimation bias with truncated mixture of continuous
  distributional quantile critics.
\newblock In \emph{International Conference on Machine Learning}, 2020.

\bibitem[Lee et~al.(2025)Lee, Hwang, Kim, Kim, Tai, Subramanian, Wurman, Choo,
  Stone, and Seno]{huang2024simba}
Hojoon Lee, Dongyoon Hwang, Donghu Kim, Hyunseung Kim, Jun~Jet Tai, Kaushik
  Subramanian, Peter~R. Wurman, Jaegul Choo, Peter Stone, and Takuma Seno.
\newblock {SimBa}: Simplicity bias for scaling up parameters in deep
  reinforcement learning.
\newblock In \emph{International Conference on Learning Representations}, 2025.

\bibitem[Liang et~al.(2023)Liang, Huang, Xia, Xu, Hausman, Ichter, Florence,
  and Zeng]{liang2023code}
Jacky Liang, Wenlong Huang, Fei Xia, Peng Xu, Karol Hausman, Brian Ichter, Pete
  Florence, and Andy Zeng.
\newblock Code as policies: Language model programs for embodied control.
\newblock In \emph{2023 IEEE International Conference on Robotics and
  Automation (ICRA)}, 2023.

\bibitem[Ma et~al.(2024)Ma, Liang, Wang, Huang, Bastani, Jayaraman, Zhu, Fan,
  and Anandkumar]{ma2024eureka}
Yecheng~Jason Ma, William Liang, Guanzhi Wang, De-An Huang, Osbert Bastani,
  Dinesh Jayaraman, Yuke Zhu, Linxi Fan, and Anima Anandkumar.
\newblock Eureka: Human-level reward design via coding large language models.
\newblock In \emph{The Twelfth International Conference on Learning
  Representations}, 2024.

\bibitem[Ma et~al.(2025)Ma, Hejna, Wahid, Fu, Shah, Liang, Xu, Kirmani, Xu,
  Driess, Xiao, Tompson, Bastani, Jayaraman, Yu, Zhang, Sadigh, and
  Xia]{arxiv_2411.04549}
Yecheng~Jason Ma, Joey Hejna, Ayzaan Wahid, Chuyuan Fu, Dhruv Shah, Jacky
  Liang, Zhuo Xu, Sean Kirmani, Peng Xu, Danny Driess, Ted Xiao, Jonathan
  Tompson, Osbert Bastani, Dinesh Jayaraman, Wenhao Yu, Tingnan Zhang, Dorsa
  Sadigh, and Fei Xia.
\newblock Vision language models are in-context value learners.
\newblock In \emph{International Conference on Learning Representations}, 2025.

\bibitem[Miech et~al.(2019)Miech, Zhukov, Alayrac, Tapaswi, Laptev, and
  Sivic]{miech2019howto100m}
Antoine Miech, Dimitri Zhukov, Jean-Baptiste Alayrac, Makarand Tapaswi, Ivan
  Laptev, and Josef Sivic.
\newblock Howto100m: Learning a text-video embedding by watching hundred
  million narrated video clips.
\newblock In \emph{Proceedings of the IEEE/CVF international conference on
  computer vision}, 2019.

\bibitem[Mildenhall et~al.(2020)Mildenhall, Srinivasan, Tancik, Barron,
  Ramamoorthi, and Ng]{mildenhall2020nerf}
Ben Mildenhall, Pratul~P. Srinivasan, Matthew Tancik, Jonathan~T. Barron, Ravi
  Ramamoorthi, and Ren Ng.
\newblock Nerf: Representing scenes as neural radiance fields for view
  synthesis.
\newblock In \emph{European Conference on Computer Vision}, 2020.

\bibitem[Moroncelli et~al.(2024)Moroncelli, Soni, Shahid, Maccarini, Forgione,
  Piga, Spahiu, and Roveda]{moroncelli2024integrating}
Angelo Moroncelli, Vishal Soni, Asad~Ali Shahid, Marco Maccarini, Marco
  Forgione, Dario Piga, Blerina Spahiu, and Loris Roveda.
\newblock Integrating reinforcement learning with foundation models for
  autonomous robotics: Methods and perspectives.
\newblock \emph{arXiv preprint arXiv:2410.16411}, 2024.

\bibitem[Nair et~al.(2022)Nair, Rajeswaran, Kumar, Finn, and
  Gupta]{nair2022r3m}
Suraj Nair, Aravind Rajeswaran, Vikash Kumar, Chelsea Finn, and Abhinav Gupta.
\newblock {R3M}: A universal visual representation for robot manipulation.
\newblock In \emph{Conference on Robot Learning}, 2022.

\bibitem[Pan et~al.(2024)Pan, Yaman, Nesti, Mallik, Allievi, Velipasalar, and
  Ren]{pan2024vlp}
Chenbin Pan, Burhaneddin Yaman, Tommaso Nesti, Abhirup Mallik, Alessandro~G
  Allievi, Senem Velipasalar, and Liu Ren.
\newblock Vlp: Vision language planning for autonomous driving.
\newblock In \emph{Proceedings of the IEEE/CVF Conference on Computer Vision
  and Pattern Recognition}, pp.\  14760--14769, 2024.

\bibitem[Patel et~al.(2023)Patel, Eghbalzadeh, Kamra, Iuzzolino, Jain, and
  Desai]{patel2023pretrained}
Dhruvesh Patel, Hamid Eghbalzadeh, Nitin Kamra, Michael~Louis Iuzzolino, Unnat
  Jain, and Ruta Desai.
\newblock Pretrained language models as visual planners for human assistance.
\newblock In \emph{Proceedings of the IEEE/CVF International Conference on
  Computer Vision}, 2023.

\bibitem[Peng et~al.(2018)Peng, Abbeel, Levine, and Van~de
  Panne]{peng2018deepmimic}
Xue~Bin Peng, Pieter Abbeel, Sergey Levine, and Michiel Van~de Panne.
\newblock Deepmimic: Example-guided deep reinforcement learning of
  physics-based character skills.
\newblock \emph{ACM Transactions on Graphics}, 37\penalty0 (4):\penalty0 143,
  2018.

\bibitem[Raffin et~al.(2021)Raffin, Hill, Gleave, Kanervisto, Ernestus, and
  Dormann]{raffin2021stable}
Antonin Raffin, Ashley Hill, Adam Gleave, Anssi Kanervisto, Maximilian
  Ernestus, and Noah Dormann.
\newblock Stable-baselines3: Reliable reinforcement learning implementations.
\newblock \emph{Journal of Machine Learning Research}, 2021.

\bibitem[Raffin et~al.(2022)Raffin, Kober, and Stulp]{raffin2022smooth}
Antonin Raffin, Jens Kober, and Freek Stulp.
\newblock Smooth exploration for robotic reinforcement learning.
\newblock In \emph{Conference on robot learning}, 2022.

\bibitem[Rocamonde et~al.(2024)Rocamonde, Montesinos, Nava, Perez, and
  Lindner]{rocamonde2024visionlanguagemodelszeroshotreward}
Juan Rocamonde, Victoriano Montesinos, Elvis Nava, Ethan Perez, and David
  Lindner.
\newblock Vision-language models are zero-shot reward models for reinforcement
  learning.
\newblock In \emph{The Twelfth International Conference on Learning
  Representations}, 2024.

\bibitem[Sferrazza et~al.(2024)Sferrazza, Huang, Lin, Lee, and
  Abbeel]{sferrazza2024humanoidbench}
Carmelo Sferrazza, Dun-Ming Huang, Xingyu Lin, Youngwoon Lee, and Pieter
  Abbeel.
\newblock Humanoidbench: Simulated humanoid benchmark for whole-body locomotion
  and manipulation, 2024.

\bibitem[Shinn et~al.(2024)Shinn, Cassano, Gopinath, Narasimhan, and
  Yao]{shinn2024reflexion}
Noah Shinn, Federico Cassano, Ashwin Gopinath, Karthik Narasimhan, and Shunyu
  Yao.
\newblock Reflexion: Language agents with verbal reinforcement learning.
\newblock \emph{Advances in Neural Information Processing Systems}, 2024.

\bibitem[Sontakke et~al.(2024)Sontakke, Zhang, Arnold, Pertsch, B{\i}y{\i}k,
  Sadigh, Finn, and Itti]{sontakke2024roboclip}
Sumedh Sontakke, Jesse Zhang, S{\'e}b Arnold, Karl Pertsch, Erdem B{\i}y{\i}k,
  Dorsa Sadigh, Chelsea Finn, and Laurent Itti.
\newblock Roboclip: One demonstration is enough to learn robot policies.
\newblock \emph{Advances in Neural Information Processing Systems}, 2024.

\bibitem[Sutton(2018)]{sutton2018reinforcement}
Richard~S Sutton.
\newblock Reinforcement learning: An introduction.
\newblock \emph{A Bradford Book}, 2018.

\bibitem[van~der Maaten \& Hinton(2008)van~der Maaten and
  Hinton]{JMLR:v9:vandermaaten08a}
Laurens van~der Maaten and Geoffrey Hinton.
\newblock Visualizing data using t-sne.
\newblock \emph{Journal of Machine Learning Research}, 2008.

\bibitem[Wang et~al.(2024{\natexlab{a}})Wang, He, Li, Li, Yu, Ma, Li, Chen,
  Chen, Wang, Luo, Liu, Wang, Wang, and Qiao]{wang2024internvid}
Yi~Wang, Yinan He, Yizhuo Li, Kunchang Li, Jiashuo Yu, Xin Ma, Xinhao Li, Guo
  Chen, Xinyuan Chen, Yaohui Wang, Ping Luo, Ziwei Liu, Yali Wang, Limin Wang,
  and Yu~Qiao.
\newblock Internvid: A large-scale video-text dataset for multimodal
  understanding and generation.
\newblock In \emph{Proceedings of the Twelfth International Conference on
  Learning Representations}, 2024{\natexlab{a}}.

\bibitem[Wang et~al.(2024{\natexlab{b}})Wang, Sun, Zhang, Xian, Biyik, Held,
  and Erickson]{wang2024rlvlmf}
Yufei Wang, Zhanyi Sun, Jesse Zhang, Zhou Xian, Erdem Biyik, David Held, and
  Zackory Erickson.
\newblock {RL-VLM-F}: Reinforcement learning from vision language foundation
  model feedback.
\newblock In \emph{International Conference on Machine Learning},
  2024{\natexlab{b}}.

\bibitem[Wo{\l}czyk et~al.(2021)Wo{\l}czyk, Zaj{\k{a}}c, Pascanu, Kuci{\'n}ski,
  and Mi{\l}o{\'s}]{wolczyk2021continual}
Maciej Wo{\l}czyk, Micha{\l} Zaj{\k{a}}c, Razvan Pascanu, {\L}ukasz
  Kuci{\'n}ski, and Piotr Mi{\l}o{\'s}.
\newblock Continual world: A robotic benchmark for continual reinforcement
  learning.
\newblock \emph{Advances in Neural Information Processing Systems}, 2021.

\bibitem[Wu et~al.(2024)Wu, Yin, Feng, He, Li, HAO, and Long]{wu2024ivideogpt}
Jialong Wu, Shaofeng Yin, Ningya Feng, Xu~He, Dong Li, Jianye HAO, and
  Mingsheng Long.
\newblock ivideo{GPT}: Interactive video{GPT}s are scalable world models.
\newblock In \emph{The Thirty-eighth Annual Conference on Neural Information
  Processing Systems}, 2024.

\bibitem[Xie et~al.(2018)Xie, Sun, Huang, Tu, and Murphy]{xie2018rethinking}
Saining Xie, Chen Sun, Jonathan Huang, Zhuowen Tu, and Kevin Murphy.
\newblock Rethinking spatiotemporal feature learning: Speed-accuracy trade-offs
  in video classification.
\newblock In \emph{Proceedings of the European conference on computer vision
  (ECCV)}, 2018.

\bibitem[Xie et~al.(2024)Xie, Zhao, Wu, Liu, Luo, Zhong, Yang, and
  Yu]{xie2024textreward}
Tianbao Xie, Siheng Zhao, Chen~Henry Wu, Yitao Liu, Qian Luo, Victor Zhong,
  Yanchao Yang, and Tao Yu.
\newblock Text2reward: Reward shaping with language models for reinforcement
  learning.
\newblock In \emph{The Twelfth International Conference on Learning
  Representations}, 2024.

\bibitem[Xu et~al.(2024)Xu, Wu, Wen, Li, Liu, Che, and Tang]{xu2024survey}
Zhiyuan Xu, Kun Wu, Junjie Wen, Jinming Li, Ning Liu, Zhengping Che, and Jian
  Tang.
\newblock A survey on robotics with foundation models: toward embodied ai.
\newblock \emph{arXiv preprint arXiv:2402.02385}, 2024.

\bibitem[Yang et~al.(2024)Yang, Tjia, Berg, Damen, Agrawal, and
  Gupta]{yang2024rank2reward}
Daniel Yang, Davin Tjia, Jacob Berg, Dima Damen, Pulkit Agrawal, and Abhishek
  Gupta.
\newblock Rank2reward: Learning shaped reward functions from passive video.
\newblock In \emph{IEEE International Conference on Robotics and Automation
  (ICRA)}, 2024.

\bibitem[Yu et~al.(2025)Yu, Zhang, Soora, Huang, Huang, Tokekar, and
  Gao]{yu2025genflowrl}
Kelin Yu, Sheng Zhang, Harshit Soora, Furong Huang, Heng Huang, Pratap Tokekar,
  and Ruohan Gao.
\newblock Genflowrl: Shaping rewards with generative object-centric flow in
  visual reinforcement learning.
\newblock In \emph{Proceedings of the IEEE/CVF International Conference on
  Computer Vision (ICCV)}, 2025.

\bibitem[Yu et~al.(2019)Yu, Quillen, He, Julian, Hausman, Finn, and
  Levine]{yu2019meta}
Tianhe Yu, Deirdre Quillen, Zhanpeng He, Ryan Julian, Karol Hausman, Chelsea
  Finn, and Sergey Levine.
\newblock Meta-world: A benchmark and evaluation for multi-task and meta
  reinforcement learning.
\newblock In \emph{Proceedings of the Third Conference on Robot Learning},
  2019.

\bibitem[Zhang et~al.(2025)Zhang, Luo, Anwar, Sontakke, Lim, Thomason, Biyik,
  and Zhang]{zhang2025rewind}
Jiahui Zhang, Yusen Luo, Abrar Anwar, Sumedh~A. Sontakke, Joseph~J. Lim, Jesse
  Thomason, Erdem Biyik, and Jesse Zhang.
\newblock {ReWiND}: Language-guided rewards teach robot policies without new
  demonstrations.
\newblock In \emph{Conference on Robot Learning}, 2025.

\end{thebibliography}
